\documentclass[letterpaper, 10 pt, conference]{ieeeconf}  
\IEEEoverridecommandlockouts                              
\usepackage{graphicx}
\usepackage{amsmath,amssymb}
\DeclareMathOperator*{\argmax}{arg\,max}
\usepackage{cite}
\usepackage{balance}
\usepackage{graphicx}
\usepackage{placeins}
\usepackage[ruled,vlined,linesnumbered]{algorithm2e}
\usepackage{algpseudocode}
\usepackage{enumerate}
\usepackage{multirow}
\usepackage{array}
\usepackage{booktabs}
\usepackage{threeparttable}
\usepackage{gensymb}
\usepackage{subfig}
\usepackage{xcolor}

\newcolumntype{L}[1]{>{\centering\arraybackslash}m{#1}}
\renewcommand{\tabcolsep}{3pt}

\newcommand{\eg}{{e.g.}\ }
\newcommand{\R}{{\mathbb{R}}}

\usepackage{caption}
\newtheorem{thm}{Theorem}[section]

\newtheorem{prob}[thm]{Problem}

\usepackage{tikz}
\usepackage{textcomp}
\usepackage{hyperref}
\usepackage{lipsum}

\newcommand\copyrighttext{%
  \footnotesize \textcopyright 2022 IEEE. Personal use of this material is permitted.
  Permission from IEEE must be obtained for all other uses, in any current or future
  media, including reprinting/republishing this material for advertising or promotional
  purposes, creating new collective works, for resale or redistribution to servers or
  lists, or reuse of any copyrighted component of this work in other works.}
\newcommand\copyrightnotice{%
\begin{tikzpicture}[remember picture,overlay]
\node[anchor=south,yshift=10pt] at (current page.south) {\fbox{\parbox{\dimexpr\textwidth-\fboxsep-\fboxrule\relax}{\copyrighttext}}};
\end{tikzpicture}%
}
\title{\LARGE \bf Autonomous Vehicle Parking in Dynamic Environments: An Integrated System with Prediction and Motion Planning 
}

\author{Jessica Leu, Yebin Wang, Masayoshi Tomizuka, and Stefano Di Cairano 
\thanks{J. Leu and M. Tomizuka are with with the Department of Mechanical Engineering, University of California, Berkeley, CA 94720, USA. This work was done while J. Leu was a research intern at Mitsubishi Electric Research Laboratories. (email: \tt\small \{jess.leu24,tomizuka\}@berkeley.edu).}  
\thanks{Y. Wang and S. Di Cairano are with Mitsubishi Electric Research Laboratories, Cambridge, MA 02139, USA (email: \tt\small \{yebinwang,dicairano\}@ieee.org).}
}

\begin{document}

\maketitle
\copyrightnotice
\thispagestyle{empty}
\pagestyle{empty}

\begin{abstract}
This paper presents an integrated motion planning system for autonomous vehicle (AV) parking in the presence of other moving vehicles. The proposed system includes 1) a hybrid environment predictor that predicts the motions of the surrounding vehicles and 2) a strategic motion planner that reacts to the predictions. The hybrid environment predictor performs short-term predictions via an extended Kalman filter and an adaptive observer. It also combines short-term predictions with a driver behavior cost-map to make long-term predictions. The strategic motion planner comprises 1) a model predictive control-based safety controller for trajectory tracking; 2) a search-based retreating planner for finding an evasion path in an emergency; 3) an optimization-based repairing planner for planning a new path when the original path is invalidated. Simulation validation demonstrates the effectiveness of the proposed method in terms of initial planning, motion prediction, safe tracking, retreating in an emergency, and trajectory repairing.
\end{abstract}
\vspace{-2.5mm}

\maketitle
\thispagestyle{empty}
\pagestyle{empty}

\section{Introduction}\label{sec:intro}

Fully autonomous parking~\cite{conner2007valet, min2013design} remains challenging especially in a dynamic environment with multiple independent agents, not only because it involves motion planning in a tight space, but also because autonomous vehicles (AVs) should intelligently react to the surrounding vehicles, i.e., obstacle vehicles (OVs). In contrast to driving on roads or highways, vehicle motions in parking areas do not have a clear set of rules to follow and largely depend on the driver's intention or even skill level. These make the environment prediction in parking challenging, hence an autonomous parking system that integrates both prediction and planning is possibly necessary.

Motion prediction is crucial because it determines the safety constraints of the planning modules and thus the feasibility and smoothness of the motion plan \cite{leu2019motion}. Particularly, an accurate short-term motion prediction enables the AV to plan and react safely to the OVs; whereas long-term plan/mode prediction allows the AV to plan more efficiently and smoothly. This work proposes a model-based hybrid predictor to perform both short-term motion and long-term mode predictions by observing the poses of OVs. A major challenge in short-term prediction is to estimate the OV's steering angle. 
We use extended Kalman filter (EKF) to reconstruct OV's velocity, and then, resort to an adaptive observer for the steering estimation. Despite of the difficulty in predicting exact long-term motions, we observe that a driver generally follows some routes as a result of driving conventions (e.g., cars should stay on their left hand side in Japan). Also, vehicles' motion throughout parking/leaving can be captured by several ``modes'' (e.g., maneuvering into/out of tight-spaces and cruising on aisles). Based on these two priors, we use a cost-map \cite{costmap08, costmap14} to capture these routes, combine the short-term predictions to determine OVs' modes, and make long-term predictions.

Motion planning for AVs is another challenge in parking scenarios. General motion planning algorithms~\cite{dolgov2008practical, mcnaughton2011motion, ma2015efficient, islam2016connect, aine2016multi, chen2017constrained} 
are not directly suitable for parking in the presence of OVs which requires rapid replanning for complicated driving maneuvers. On the other hand, motion planners specialized in autonomous parking either fail to integrate short-horizon planning with long-horizon planning or cannot incorporate online path repairing upon new obstacles in dynamic environments~\cite{hsieh2008parking, kummerle2009autonomous, han2011unified, min2013design, klemm2016autonomous, tazaki2017parking}. 
A scenario-aware planner that implements multiple strategies can be effective in terms of computation time, leading to a high replanning rate, and possibly safety guarantees. In this work, we first generate a long-term motion reference with Bi-Directional A-Search Guided Tree (BIAGT)~\cite{wang2019improved}. Then, a strategic motion planner, based on the results of the hybrid environment predictor, implements three strategies: 1) model predictive control (MPC)-based safety controller \cite{rawlings2017model} for trajectory tracking if the reference remains valid regarding the environment, 2) search-based retreat-planning that quickly finds an evasion path in an emergency, and 3) optimization-based repair-planning when the reference is invalidated. 

This work presents an integrated system for autonomous vehicle parking in dynamic environments. 
Main contributions are threefold as follows:
\begin{itemize}
  \item A model-based hybrid environment predictor predicts short-term motions and long-term modes.
  \item A strategic motion planner is presented to efficiently plan under different situations.
  \item Simulation is performed to show the effectiveness of the proposed system. 
\end{itemize}


\section{Related Works} \label{sec:RelatedWork}
\subsection{Predictor}
    Research in vehicle motion prediction has attracted a lot of interests, and results in numerous contributions, e.g., short-term motion prediction methods \cite{schubert2008comparison, houenou2013vehicle, 4895669} and long-term plans/modes prediction methods \cite{houenou2013vehicle, 4895669,shen2020collision, 2015classify,shen2020parkpredict, deo2018would, 6025208}. 
    It is arguably true that most research in this area is related to road driving. Interested readers are referred to an extensive survey in~\cite{lefevre2014survey}. 
    In contrast, vehicle motion prediction in parking is less explored. In~\cite{jeong2018sampling}, an interacting multiple model (IMM) filter is used to predict short-term trajectories in parking. Focusing on long-term prediction, \cite{govea2004moving} first trains a trajectory cluster classifier, and then acquires the mean-value trajectory of the classified cluster. In~\cite{shen2020parkpredict}, the classified driver's intent and the vehicle's pose history are used to generate short-term motion predictions with a Long Short-Term Memory network. Purely data driven methods are not desirable for two reasons: 1) the lack of guarantees; 2) their performance depends largely on the training data set, and they may present a larger prediction error if the data set is chosen poorly. Prediction network over fitting may also be a concern. To the best of our knowledge, there are not extensive studies on a predictor fusing both short-term and long-term predictions for parking. 

\subsection{Planning}
Prevailing motion planning approaches fall into three categories: search-based~\cite{HarNilRap68,Ste93,LikFerGor05}, sampling-based~\cite{kavraki1996probabilistic, lavalle2001randomized, karaman2011sampling} and optimization-based~\cite{zucker2013chomp, schulman2014motion, gutjahr2016lateral, zhang2018autonomous}. 
Sampling-based planners could raise concerns in risk-sensitive tasks due to their non-deterministic nature, while optimization-based planners are only locally optimal and often need to work with global planners~\cite{zucker2010optimization, xu2012real, dai2018improving, zhang2018autonomous,leu2021efficient}. Various search-based motion planners are widely adopted by autonomous vehicles for their computation efficiency with well-chosen motion primitives and heuristics~\cite{UrmAnhBag08, dolgov2008practical, montemerlo2008junior, dolgov2010path, chen2015kinodynamic, ajanovic2018search, chen2015path, klaudt2017priori}. 

If the parking environment changes, the initial long-term trajectory may need to be repaired. Real-time trajectory repairing methods include online heuristic update~\cite{koenig2002d, sun2010moving, oral2015mod}, pruning and reconnecting sampling-based search structures~\cite{ferguson2006replanning, chandler2017online, adiyatov2017novel, qi2020mod}, and spline-based kinodynamic search~\cite{ding2018trajectory, ding2019efficient}. The heuristic update method is not directly applicable to our tree-based search structure in BIAGT and pruning is less efficient for parking scenarios. Spline-based kinodynamic search is relatively efficient but the original trajectory is not utilized. On the other hand, we observe that alternative feasible solutions in the repairing scenarios are often in the same homotopy class as the original trajectory. Therefore, optimization-based methods~\cite{9062306,schulman2014motion,ding2018trajectory} become suitable candidates for repairing an existing path. 

\subsection{Parking system}
As for system-level strategies for AV parking, \cite{9183957} and \cite{2020parkingsystem} present autonomous parking systems to park in static environments. In \cite{jeong2018sampling}, IMM is used for prediction with a sampling-based method for planning. The method in \cite{shen2020collision} first predicts the strategy of OVs, and then selects the navigation strategy of the ego AV. However, the AV only traverses the roads of the parking lot but do not perform parking maneuvers. Instead, our work aims at the more complete approach of an integrated parking system that makes both short-term and long-term predictions of the environment and utilize them in the strategic planning for parking.         

\section{Methods} \label{sec:mr}

\subsection{Problem Statement}
\begin{figure}
	\centering
    \includegraphics[width=0.4\linewidth]{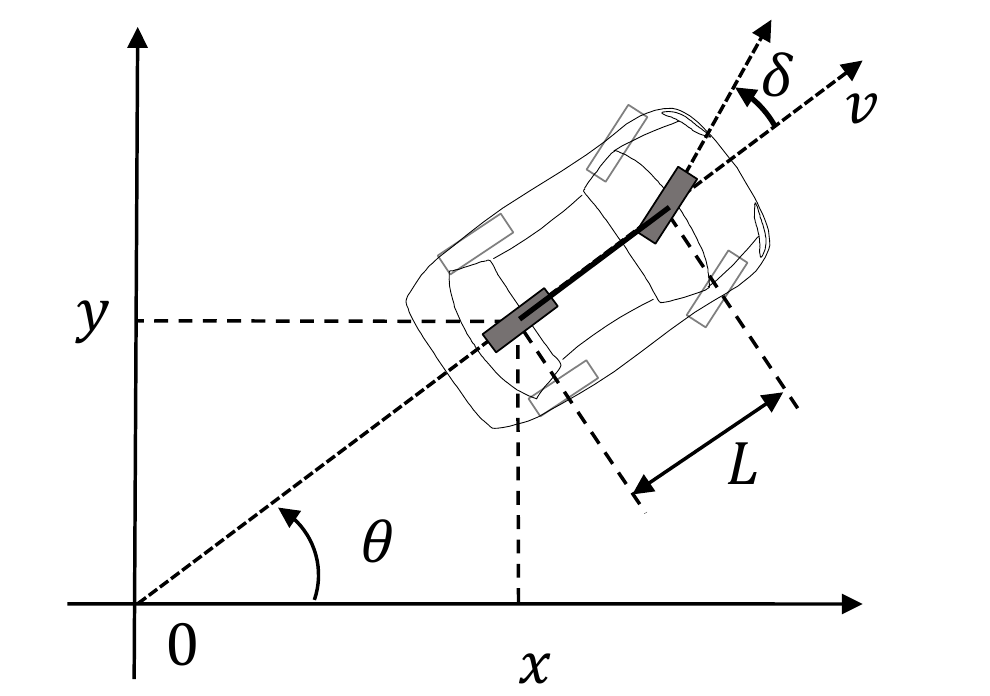}
	\caption{The simplified bicycle vehicle model. $L$ is the distance between the axis of the rear wheels and the axis of the front wheels.}
	\label{fig:car}
\end{figure}

Consider the planning problem with vehicle dynamics:
\begin{equation}\label{eq:plant}\small
\dot X = f(X) + g(X,u),
\end{equation}
\noindent where $X = [x, y, \theta]^\top$ denotes the 2D coordinates and the vehicle heading, and $u = [\delta, v]^\top$ is the control input that includes longitudinal velocity and steering angle. A collision-free configuration space $\mathcal C_{free} \subset \R^{n_c}$ is the set of configurations at which the vehicle has no intersection with the obstacles. The motion planning problem considered in this paper is defined as follows:

\begin{prob}\label{prob:path_planning}
Given an initial configuration $X_0 \in \mathcal C_{free}$, a goal configuration $X_f \in \mathcal C_{free}$, and system \eqref{eq:plant}, find a feasible trajectory $\mathcal P_t$ which
\begin{enumerate}[(I)]
\item starts at $X_0$ and ends at $X_f$, while satisfying \eqref{eq:plant}; and
\item lies in the collision-free configuration space $\mathcal C_{free}$.
\end{enumerate}
\end{prob}

We use the bicycle model, illustrated by Fig.~\ref{fig:car}, to represent the vehicle motion. The discrete-time model is obtained through Euler discretization as follows:
\begin{equation}\label{model:bicycle}\small
\begin{aligned}
\begin{bmatrix} x_{k+1} \\ y_{k+1}\\ \theta_{k+1} \end{bmatrix}=  
\begin{bmatrix} x_k \\ y_k \\ \theta_{k}  \end{bmatrix}+
\begin{bmatrix} v_{k}T_s\cos(\theta_{k}) \\ v_{k}T_s\sin(\theta_{k}) \\ v_kT_s\frac{\tan{\delta_k}}{L} \end{bmatrix},
\end{aligned}
\end{equation}
where $T_s$ is the sampling time.

\subsection{Proposed Architecture}
\begin{figure}
	\centering
    \includegraphics[width=0.9\linewidth]{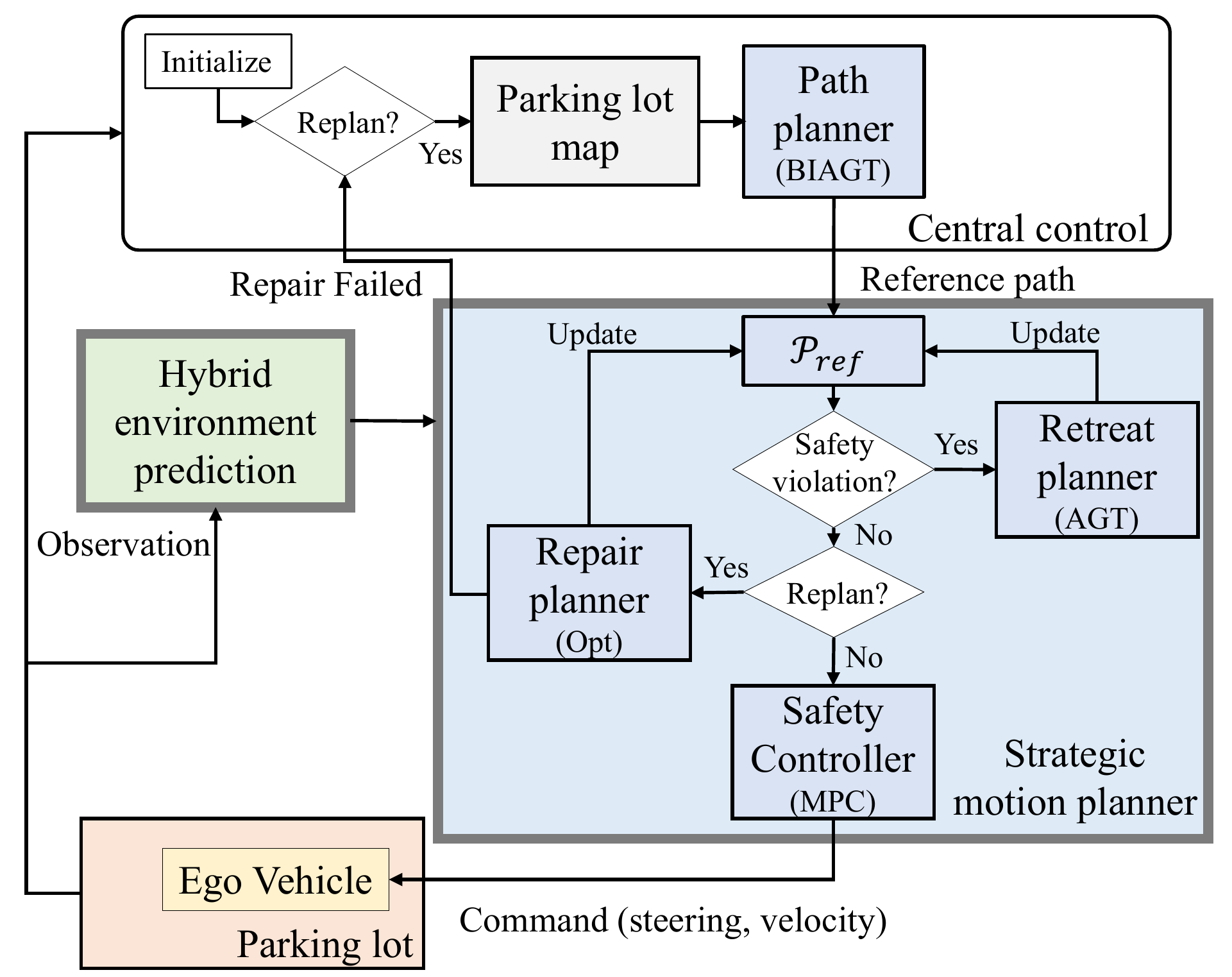}
	\caption{The integrated prediction and planning system.}
	\label{fig:system}
\end{figure}
Fig.~\ref{fig:system} shows the architecture of the proposed system. The two main components are the hybrid environment predictor (Section~\ref{sec:predictor}) and the strategic motion planner~(Section~\ref{sec:planner}). During run-time, the central control first processes the parking lot map $M_{map}$ and generates an initial long-term trajectory $\mathcal P_{ref}$. We use BIAGT since it is guaranteed to plan a trajectory that brings the ego AV exactly to the goal, an important feature for  parking in a tight space. The hybrid environment predictor monitors the environment and predicts the movements of the OVs. Based on the prediction, the strategic motion planner first checks if the ego AV is violating the safety margin. If not, it checks if $\mathcal P_{ref}$ needs to be repaired due to the OV. If any of these situations occurs, $\mathcal P_{ref}$ will be updated. Finally, an MPC-based safety controller plans a collision free motion that tracks the latest $\mathcal P_{ref}$ in the dynamic environment. If the repair-planner cannot succeed, it requests the central control to update the map and regenerate a reference trajectory. 

\subsection{Hybrid Environment Predictor}\label{sec:predictor}
The hybrid environment predictor (summarized in Alg.~\ref{algorithm:Predictor}) contains three main parts: motion estimation, motion prediction, and mode estimation of OVs. 

\subsubsection{Cascaded motion estimation}
\begin{figure}
	\centering
    \includegraphics[width=1\linewidth]{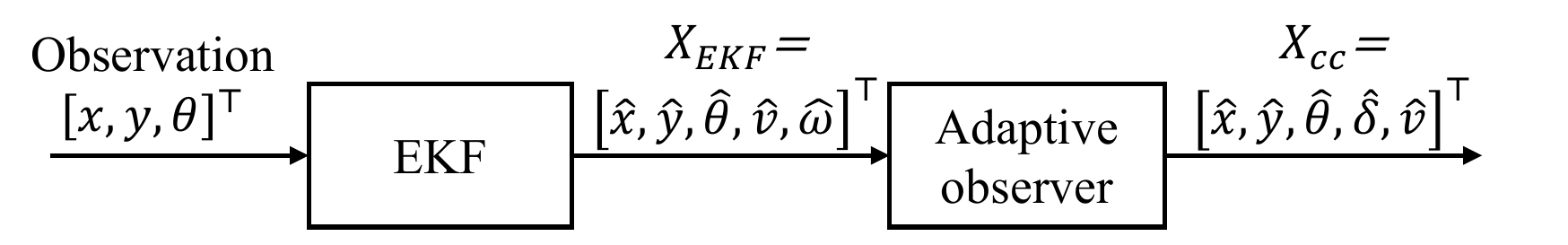}
	\caption{The architecture of the cascade motion estimator.}
	\label{fig:estimator}
\end{figure}

Motion estimation has been studied in, \eg \cite{shen2020parkpredict,jeong2018sampling}, where the state $X$ is reconstructed from the measurement of $(x,y)$ based on the unicycle model. Such a treatment is not sufficient for vehicle parking, where frequent changes of moving direction and steering actions are involved. To accurately predict the short-term motion of the OV, it is advantageous to reconstruct the control input $u$. One can either pose it as an unknown input estimation problem \cite{HouPat98} or augment the OV's system state with the control input and solve a state estimation problem. Assume that the OV motion evolves according to the model~\eqref{model:bicycle}. 
We obtain the augmented model of the OV by assuming that control input $(\delta, v)$ are piecewise constant, and estimate the augment state $[x,y,\theta,\delta,v]^\top$. 


Given the nonlinear augmented model, it is natural to apply well-established nonlinear state estimators such as EKF or particle filter for state estimation. We observe that it is not straightforward to tune EKF for accurate estimation of the steering angle. This is partially attributed to the term $v\tan(\delta)$ which involves the multiplication of unmeasured states. Meanwhile, the heavy computation presents a hurdle for the adoption of particle filter.  

A cascaded motion estimator (Fig.~\ref{fig:estimator}) is proposed to estimate the OV motion. The velocity estimation resorts to EKF and is based on the following discrete-time model:
\begin{equation}\label{model:unicycle}\small
\begin{aligned}
\begin{bmatrix} x_{k+1} \\ y_{k+1}\\ \theta_{k+1} \\ v_{k+1}\\\omega_{k+1} \end{bmatrix} & =  
\begin{bmatrix} x_k+v_{k}T_s\cos(\theta_{k})  \\ y_k+v_{k}T_s\sin(\theta_{k})  \\ \theta_{k}+T_s\omega_k \\ v_k \\ \omega_{k} \end{bmatrix} + q_k,\; q_k \sim \mathcal{N}(0,Q), \\
z_k & = X_{OV,k} + r_k,
\end{aligned}
\end{equation}
where $\omega$ is the yaw rate, $q$ is the disturbance, $z_k = [ z_{x,k}, z_{y,k}, z_{\theta,k}]^\top$ is the measurement, $X_{OV,k}=[ x_k, y_k, \theta_k]^\top$, and $r_k \sim \mathcal{N}(0,R)$. The EKF outputs $X_{EKF,k} = [\hat{x}_k, \hat{y}_k, \hat{\theta}_k, \hat{v}_k, \hat{\omega}_k]^\top$. 
The estimation of the steering action is based on the following model
\begin{equation}\label{adaptation}\small
\begin{aligned}
\omega = vs, \quad z_\theta = \theta,
\end{aligned}
\end{equation}
where $z_\theta$ is the measurement and $s=\frac{\tan{\delta}}{L}$. 
We construct an adaptive observer as follows:
\begin{equation}\label{adaptation}\small
\begin{aligned}
M_{k+1} & = M_{k} -T_s(gM_k+\hat{v}_k),\\
\hat{s}_{k+1} & = \hat{s}_{k} + T_s(\lambda M_k(z_\theta-\hat{\theta}_{a,k})),\\
\hat{\theta}_{a,k+1} & =  \hat{\theta}_{a,k} + T_s(\hat{s}_k\hat{v}_k +g(z_\theta-\hat{\theta}_{a,k})+M_k(\lambda M_k(z_\theta-\hat{\theta}_{a,k})),\\
\end{aligned}
\end{equation}
where $M$ is an auxiliary signal, $g, \lambda \in \mathcal{R}$ are observer gains, $\hat{\theta}_a$ is the estimated heading angle, and $\hat{v}$ is the velocity estimated by EKF. The estimated steering angle can be calculated with $\hat{s}$, i.e., $\hat{\delta} = \tan ^{ - 1} (\hat{s}L)$. It is not hard to verify that as long as $\hat v \rightarrow v$ and $v$ is non-zero, the steering angle estimate is guaranteed to converge to its true value as $t \rightarrow \infty$. The output of the cascade motion estimator is denoted as $X_{cc,k} = [\hat{x}_k, \hat{y}_k, \hat{\theta}_k, \hat{\delta}_k,\hat{v}_k]^\top$.


\subsubsection{Short-term motion prediction}
For the sake of computation efficiency, we assume the short-term motion of an OV is fully captured by the mean value of the state $X_{cc}$, and its covariance. For the mean value, we propagate the estimated states $X_{cc,k}$ forward and obtain a short-term prediction $\mathbf{X}_{H,k}= [X_{1,k}^\top,\dots, X_{H,k}^\top]^\top$ for the future $H$ steps of the time horizon. Similarly, forward propagation is carried out to obtain the covariance matrices $Pm_{H,k} = \{Pm_{k+1},\dots,Pm_{k+H}\}$ according to EKF's forward prediction formula. 
These information will facilitate long-term prediction and be used to determine the safety margin for each future time step. 

\subsubsection{Long-term mode prediction}
\begin{figure}
	\centering
    \includegraphics[width=0.8\linewidth]{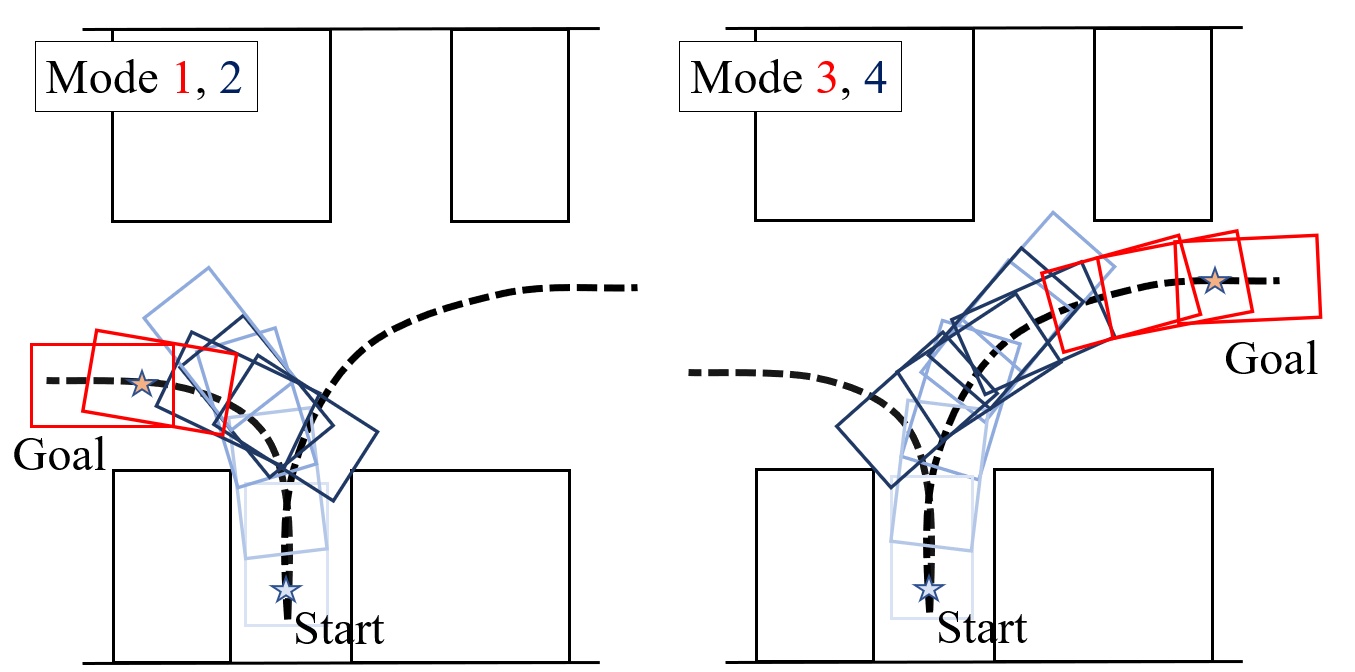}
	\caption{There are 2 routes (black dashed lines) and 2 main modes (red for ``cruising'' and blue for ``maneuvering'') in this example, resulting in a total of 4 modes for the OV: 1) cruise(exit) left;  2) maneuver left; 3) cruise(exit) right; 4) maneuver right.}
	\label{fig:mode}
\end{figure}
OV's long-term motion is dependent on the history of its state, the dynamic model, and its relative movement against the environment, where the first two factors are captured to some extent by the short-term motion prediction. 
In order to exploit the relative movement against the environment, \cite{costmap08} and \cite{costmap14} introduced a cost map to capture an OV's possible long-term movements. We adopt the same idea and construct a cost map, $M_{route}$ using a route planner~\cite{dailong}, where the cost map contains possible routes that the OV will take~\eqref{eq:costmap}. 
Also, we recognize that a vehicle in the parking lot normally runs in two modes, ``maneuvering'' and ``cruising''. Vehicles in maneuvering mode change the steering frequently and deviate from the routes (black dashed line in Fig.~\ref{fig:mode}) in the cost map in order to park or leave the narrow parking spot. Vehicles in cruising mode have small or steady steering angles and generally follow one of the routes. Vehicles are in this mode when they first enter the parking lot and are approaching a parking spot or when they got out of the parking spot and are leaving the parking lot. Including the route information, an OV that has $n$ routes to follow will have $2n$ possible modes (Fig.~\ref{fig:mode}). To determine the mode $m$ at time step $k$, i.e., $m_k$, Bayesian framework is employed to keep track of the belief of each mode, i.e., $b(m_k)$. The process is described in Alg.~\ref{algorithm:Predictor}, lines 5$\sim$8. We perform the prior believe update $p(m_k)= T_b b(m_{k-1})$ based on the previous belief, $b(m_{k-1})$. The posterior $p(m_{k}|X_{cc,k},\mathbf{X}_{H,k})$ is proportional to the prior multiplied with the conditional probability of the motion estimation and prediction given the mode, i.e., $p(m_k)p(X_{cc,k},\mathbf{X}_{H,k}|m_{k})$. The Boltzmann policy is one common way to design this conditional probability \cite{cpb}, i.e., $p(X_{cc,k},\mathbf{X}_{H,k}|m_{k}) \propto \exp(-M_{route}(m_{k},X_{cc,k},\mathbf{X}_{H,k}))f(m_k,X_{cc,k})$. The function $M_{route}(m_{k},X_{cc,k},\mathbf{X}_{H,k})$ compares the OV states and predictions with the routes:
\begin{equation}\label{eq:costmap}\small
\begin{aligned}
M_{route}(m_{k},X_{cc,k},\mathbf{X}_{H,k}) = &\min_i \|X_{m_k,i} - X_{cc,k}\|^2_{W_1} \\
&+\sum_h \min_i \|X_{m_k,i} - X_{h,k}\|^2_{W_2},
\end{aligned}
\end{equation}
where $X_{m_k,i}$ is the $ith$ waypoint of the route in mode $m_k = m$, $m \in \{1,\dots,2n\}$, $\|\mathrm{v}\|^2_W = \mathrm{v}^\top W\mathrm{v}$, and $W_{1}$ and $W_{2}$ are weighting matrices. The function $f(m_k,X_{cc,k})$ is proportional to the magnitude of the OV's steering angle and the deviation of the OV's heading angle from the final heading angle of the route. Finally, we normalize $p(m_{k}|X_{cc,k},\mathbf{X}_{H,k})$ to obtain $b(m_{k})$ and take the value of $m_k$ with the largest belief to be $\hat{m}_k$.

\subsubsection{Safety margin and safety bound}\label{sec:safety}
\begin{figure}
	\centering
    \includegraphics[width=0.8\linewidth]{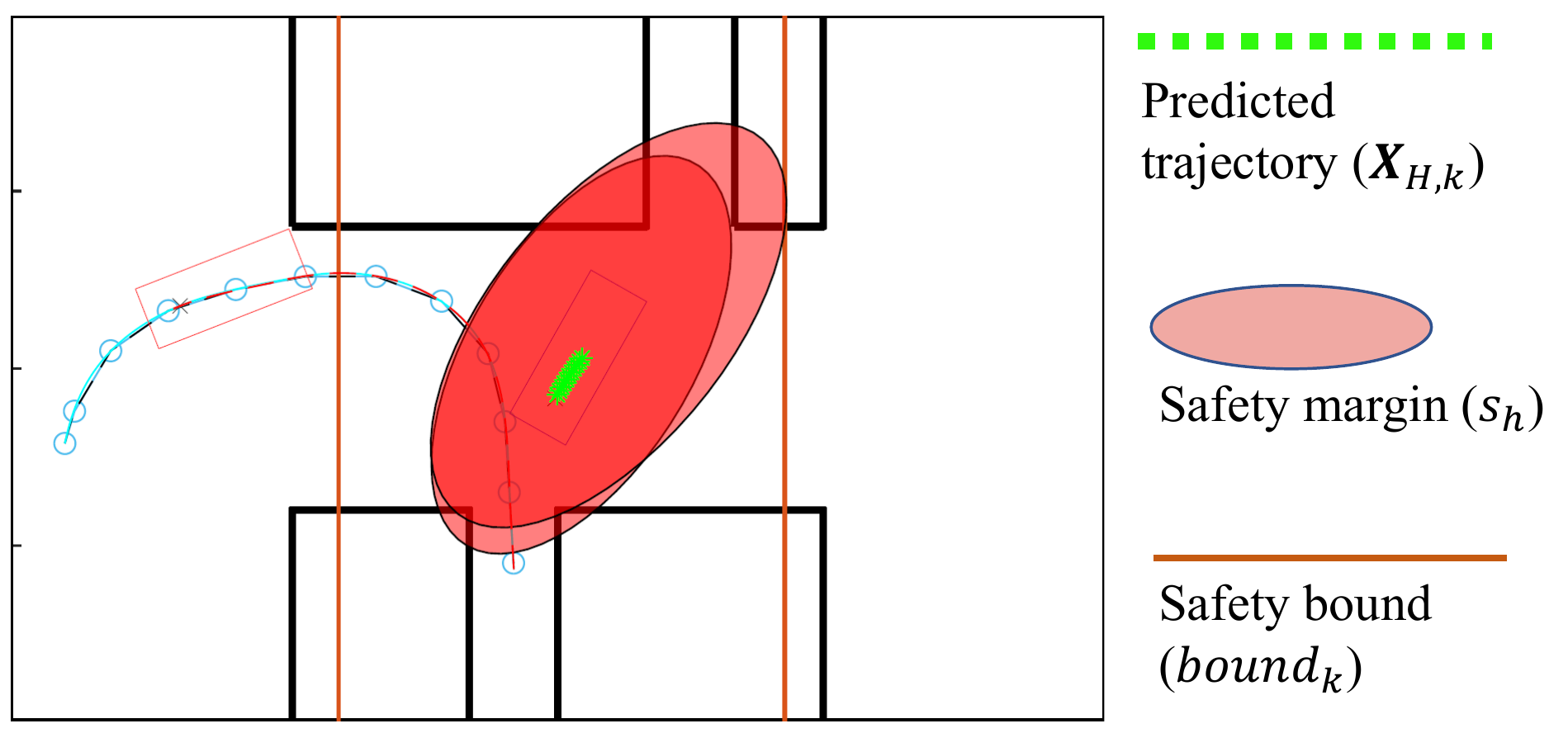}
	\caption{The hybrid predictor predicts a short-term OV trajectory (green line) and use it with mode prediction to generate the safety margins for $h=1$ and $h=H$ and the safety bound.}
	\label{fig:safety}
\end{figure}
When the OV is in cruising mode, the predictor calculates the safety margins $\mathbf{s}_{H,k}=[s_{k+1}^\top, \dots, s_{k+H}^\top]^\top$ (red areas in Fig.~\ref{fig:safety}) according to Alg.~\ref{algorithm:Predictor}, lines 9 and 10. The safety margin of the $h$th future time step is an ellipsoid and the length of the principal semi-axes $s_{k+h}$ $(h = 1,\dots,H)$ are proportional to the differential entropy of the mode belief and the covariance from the motion estimation, i.e., $s_{k+h} \propto (b_{uncertainty} C^\top Pm_{k+h}C)$. Since the movements of the OV in maneuver mode (colored in blue in Fig.~\ref{fig:mode}) are hard to predict, the predictor generates a safety bound (the bound of a convex hall of the OV's pose history, orange lines in Fig.~\ref{fig:safety}) so that the planner behaves more conservatively and keeps the ego AV away from the hardly predictable OVs. Notice that the safety margin and safety bound can also be applied to other moving obstacles such as pedestrians or motorbikes given their kinematic models and routes information.   
\begin{algorithm}[t] \small
    \bf{input} {$z_k, M_{route}$}\\
    $X_{EKF,k} \gets \tt{EKF_{unicycle}}$$(z_k)$\\
    $X_{cc,k} \gets \tt{steeringEST}$$(X_{EKF,k},z_k)$\\
    $\mathbf{X}_{H,k},Pm_{H} \gets$ $\tt{propagate}$$(X_{cc,k})$\\
    $p(m_{k}) \gets T_pb(m_{k-1})$ \\
    $p(m_{k}|X_{cc,k},\mathbf{X}_{H,k})	\propto p(X_{cc,k},\mathbf{X}_{H,k}|m_{k})p(m_{k})$\\
    $b(m_{k}) \gets$ $\tt{normalize}$$( p(m_{k}|X_{cc,k},\mathbf{X}_{H,k}))$\\
    $\hat{m}_k \gets  \argmax_{m_k} b(m_{k})$\\
    $b_{uncertainty} \propto -\sum(b(m_{k})\log(b(m_{k})))$\\
    $\mathbf{s}_{H,k} \gets$ $\tt{getSafetyMargin}$$(b_{uncertainty},Pm_{H,k})$\\
    \If{$m$ \bf{is} \tt{maneuver}}{
             $bound_k \gets$ $\tt{getConvexHall}$$(X_{cc,k},bound_{k-1})$\\
        }
    \Return $X_{cc,k},\mathbf{X}_{H,k},\mathbf{s}_{H,k},b(m_k),\hat{m}_k,bound_k$  
	\caption{Hybrid Environment Predictor}
	\label{algorithm:Predictor}
\end{algorithm}

\subsection{Strategic Motion Planner}\label{sec:planner}
With the reference trajectory $\mathcal P_{ref}$, the strategic motion planner runs the main module, the MPC-based safety controller, and two supporting modules: the retreat planner and the repair planner (Fig.~\ref{fig:system}). Both planners are activated if the ego AV's current location and reference trajectory is invalidated by the OV's movements, respectively. The strategic motion planner is summarized in Alg.~\ref{algorithm:planner}.

\subsubsection{MPC-based safety controller}
The safety controller tracks the reference trajectory $\mathcal P_{ref}$ given the safety margin and the safety bound. With $\mathcal P_{ref}$, we use an optimization-based planner to compute tracking motions in an MPC framework. Let $\mathbf{X}_{ref,k}$ be the segment of $\mathcal P_{ref}$ to track at time step $k$. $\mathbf{X}_{ref,k}$ is selected and trimmed so that it will not violate the safety margin (in all modes) nor the safety bound (in  ``maneuver'' modes). The trajectory tracking problem is formulated as follows:
\begin{prob}\label{prob:opt_planning}
Given the planning horizon $H$, the vehicle model \eqref{model:bicycle}, and the reference segment $\mathbf{X}_{ref,k}$, the optimization planner solves the problem 
\begin{equation}\label{P:MP}
\begin{aligned}
\mathbf{u}_k^* = \arg\min_{\mathbf{u}_k}\quad  &\| F(X_k) + G(X_k,\mathbf{X}_k,\mathbf{u}_k) - \mathbf{X}_{ref,k}\|^2_{W_3},\\
s.t.\quad & \mathbf{X}_k = F(X_k) + G(X_k,\mathbf{X}_k,\mathbf{u}_k),\\
\quad & (\mathbf{X}_k,\mathbf{u}_k) \in\Gamma_k,
\end{aligned}
\end{equation}
\end{prob}
where $\mathbf{X}_k= [X_{k+1}^\top, X_{k+2}^\top, \dots, X_{k+H}^\top]^\top$, $\mathbf{u}_k= [u_k^\top, u_{k+2}^\top, \dots, u_{k+H-1}^\top]^\top$, and $\Gamma_k$ defines the feasible set, $\Gamma_k = \{ (\mathbf{X}_k,\mathbf{u}_k) : X_{k+h} \in \mathcal C_{free,k},\; u_{k+h-1} \in [-u_{max}, u_{max}], \forall h = 1,\dots,H\}$.

Problem \ref{prob:opt_planning} can be readily formulated as a non-convex optimization problem using various software tools, e.g., CasADi~\cite{AndGilHor19}, and solved using nonlinear programming solvers, e.g., IPOPT. With the reference path serving as a warm start, the average solving time is around $0.06$ second. Details are omitted here. 

\subsubsection{Retreat planner} \label{sec:retreating}
\begin{figure}
 \centering        
    \begin{tabular}{@{}cc@{}}
   \begin{minipage}{.22\textwidth}
    \includegraphics[width=\textwidth]{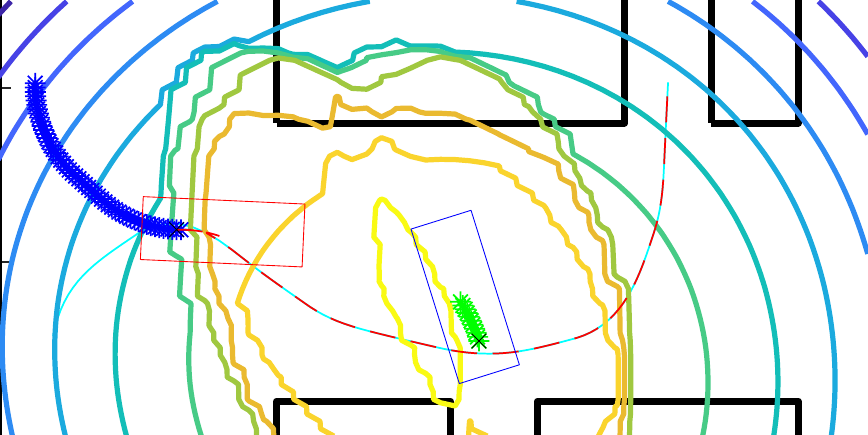}
   \end{minipage} &
    \begin{minipage}{.22\textwidth}
    \includegraphics[width=\textwidth]{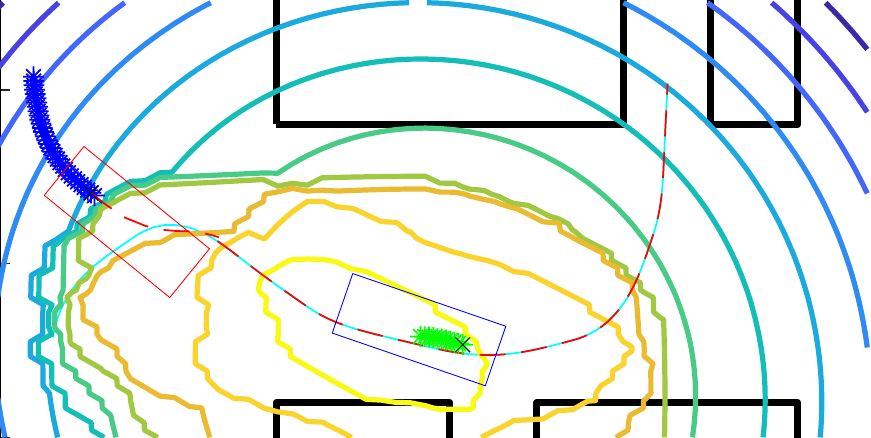}
   \end{minipage} 
  \end{tabular}
        \caption{The ego AV following the retreating plans (blue star-lines). The light-blue lines illustrate the original trajectory, the red dashed-lines combining the blue star-lines will be the new reference $\mathcal P_{ref}$, and the collision field is illustrated by the contours.}
  \label{fig:retreat}
\end{figure}
The retreat planner deals with scenarios when stopping or staying on the original reference is deemed unsafe. This can happen when the OV drives toward the ego AV, and its motion largely differs from the previous prediction - possibly violating the safety margin and causing a safety threat. Therefore, the ego AV needs to find a path and retreat from the emergency. The retreating movement is not a standard navigation problem because the ego AV hasn't had a safe goal. Instead, it needs to explore the environment to find the best goal, and thus we propose a search-based retreat planner which explores the space and quickly finds a retreating trajectory. 

As a variant of A*-based algorithms, the retreat planner constructs a tree $\mathcal T=(\mathcal V,\mathcal E)$ composed of a node set $\mathcal V \subset \mathcal C_{free}$ and an edge set $\mathcal E$, where $E(X_i,X_j) \in \mathcal E$ represents a feasible short path between $X_i$ and $X_j$. $\mathcal C_{free}$ is implicitly obtained by checking collisions with obstacles in the parking lot map $M_{map}$. Let $\mathcal M$ denote a finite set of motion primitives pre-computed through available control actions, and $V_{max}$ denotes the maximum number of nodes allowed. The retreat planner constructs a tree $\mathcal T$ from $X_{k}$ (the configuration when the retreat planning starts) and expands it according to a cost function $ \mathcal F(\cdot)$ which sums up the heuristic value $h(\cdot)$ and the arrival cost $g(\cdot)$. The heuristic~(\ref{eq:field}) is calculated based on a collision field as shown in Fig.~\ref{fig:retreat}. The field is a weighted sum of Gaussian distributions centered at waypoints of both the predicted trajectory $\mathbf{X}_{H,k}$, i.e., $X_{h,k}$, and the routes on the cost map, i.e., $X_{m_k,i}$, and
\begin{equation}\label{eq:field}\small
h(X) = \sum_{m_k,i} b(m_k) e^{-\| X - X_{m_k,i} \|^2_{W_4}} + \sum_{h} ce^{-\| X -  X_{h,k}\|^2_{W_5}},
\end{equation}
where $c$ is a weighting constant.
The planning is completed if the ego AV finds a trajectory that keeps it away from the OV at a safe clearance. If the number of nodes in $\mathcal T$ reaches $V_{\max}$, the trajectory giving the maximum clearance is chosen.

\subsubsection{Trajectory repair planner}  \label{sec:replan}
\begin{figure}
 \centering        
    \begin{tabular}{@{}cc@{}}
   \begin{minipage}{.22\textwidth}
    \includegraphics[width=\textwidth]{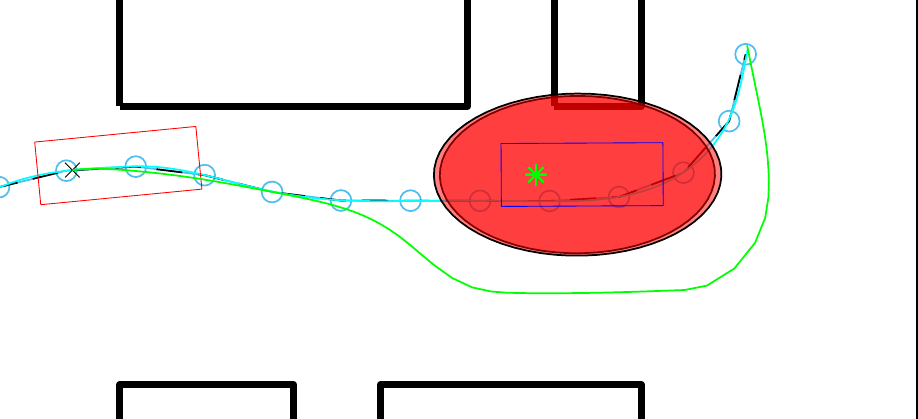}
   \end{minipage} &
    \begin{minipage}{.22\textwidth}
    \includegraphics[width=\textwidth]{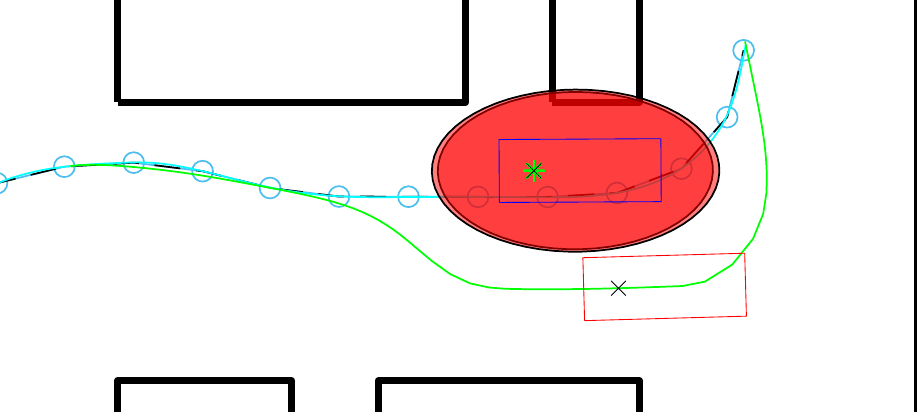}
   \end{minipage} 
  \end{tabular}
        \caption{The ego AV (red vehicle) following the repaired plan (green lines) that is calculated from the blocked original trajectory (light-blue lines).}
  \label{fig:repairing}
\end{figure}
OV's movements could invalidate the ego AV's reference trajectory. Fig.~\ref{fig:repairing} illustrates one such case, where an OV, represented by the blue box, stops on the AV's reference trajectory, represented as the light-blue line. The safety controller will command the ego AV to stop on the reference trajectory when the area in front is infeasible. Unless receiving a new reference trajectory, the safety controller will stop the ego AV and wait for the OV to clear - not efficient if the OV stops for a long time. It is reasonable to update $\mathcal P_{ref}$, so that the safety planner can command the ego AV to go around the OV and merge back to the original path. We notice that the repaired trajectory usually lies in the same homotopy class as the original one, which makes an optimization-based repairing strategy a viable solution. To obtain a repaired path quickly, we conduct repair planning over 2D space, i.e., $X_{repair}=[x,y]^\top$, and modify the constraints accordingly. It is understood that the resultant path, despite being collision-free, cannot always be followed accurately, causing the AV to collide into obstacles. We therefore verify the path and accept the repaired trajectory (as shown in Fig.~\ref{fig:repairing}) only if it passes the kinematic feasibility check. 
If the repairing fails, the central control will be notified to take over the repairing task.

\begin{algorithm}[t] \small
    \bf{input} {$M_{map},M_{route},X_{goal}$} \\
    $\tt{RequestCentralPathPlanning}$$(X_{goal})$\\
	\While {$X_{goal}$ \bf{not reached}}{
        \If{$\tt{ReceiveCentralMsg}$}{
                    $\mathcal P_{ref} \gets$ $\tt{Update}$$(\mathcal P_{ref})$\\
        }
	    $z_k \gets$ $\tt{GetMeasurements}$ \\ 
    	$X_{cc,k},\mathbf{X}_{H,k},\mathbf{s}_{H,k},b(m_k),\hat{m}_k,bound_k \gets$ $\tt{ENVPredict}$$(z_k,M_{route})$ \\ 
        $flag_{retreat} \gets$ $\tt{SafetyCheck}$$(X_{cc,k}, s_k,bound_k,\hat{m}_k)$ \\ 
        \If{$flag_{retreat}$}{
            $\mathcal P_{ref} \gets$ $\tt{PlanRetreat}$$(M_{map},M_{route},X_{cc,k},\mathbf{X}_{H,k},b(m_k))$\\
        }
        $\mathbf{X}_{OV,history} \gets$ $\tt{OV\_MotionHistory}$$(X_{cc,k})$
        $flag_{repair} \gets$ $\tt{BlockerCheck}$$(\mathbf{X}_{OV,history}, \mathcal P_{ref})$ \\
        \If{$flag_{repair}$}{
                $\mathcal P_{ref},flag_{fail}  \gets$ $\tt{PlanRepair}$$( P_{ref},X_{cc,k})$\\
        }
        \If{$flag_{fail}$}{
                $\tt{RequestCentralPathPlanning}$$(X_{goal},X_{cc,k})$\\
        }
        $\mathbf{X}_{ref,k} \gets$ $\tt{setXref}$$(\mathcal P_{ref},\mathbf{X}_{H,k},\mathbf{s}_{H,k},\hat{m}_k,bound_k)$\\
        $u_{k} \gets$ $\tt{SafetyController}$$(\mathbf{X}_{ref,k},\mathbf{X}_{H,k},\mathbf{s}_{H,k})$\\
	}
    \Return $u_k$
	\caption{The Strategic Motion Planner}
	\label{algorithm:planner}
\end{algorithm}


\section{Empirical Evaluation} \label{sec:sim}
\begin{figure*}[t]
 \centering
   \begin{tabular}{@{}cccc@{}}
   \begin{minipage}{.23\textwidth}
    \includegraphics[width=\textwidth]{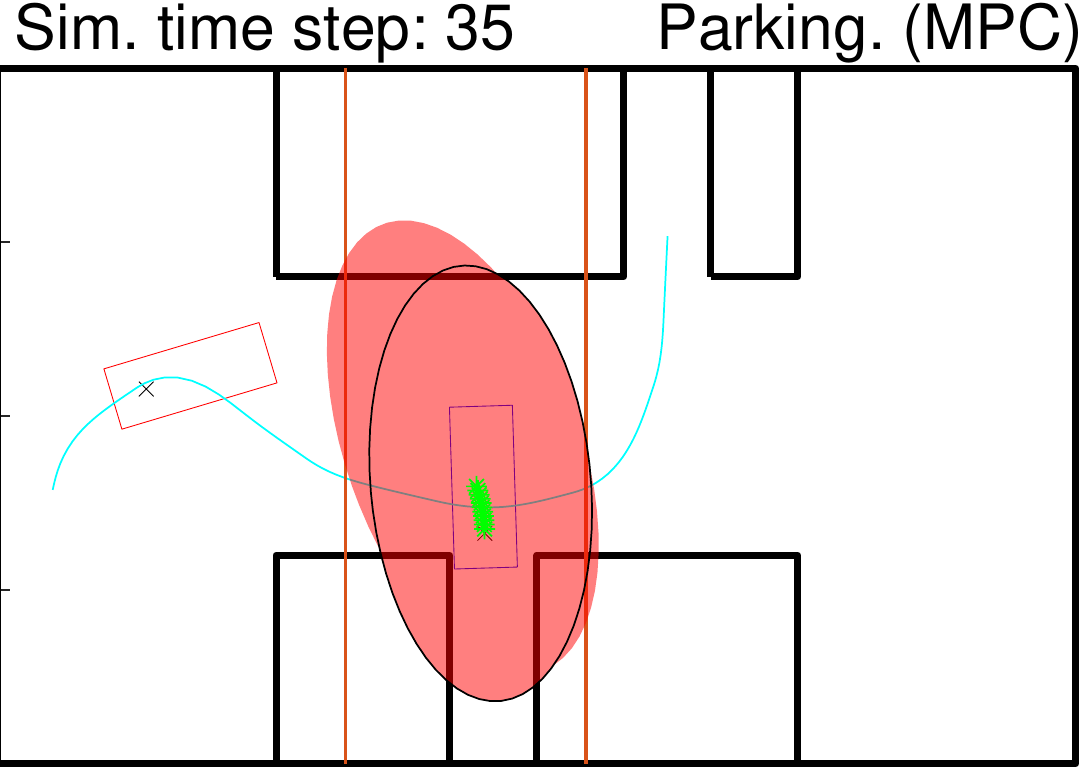}
     \captionof*{figure}{(a)}
   \end{minipage} &
    \begin{minipage}{.23\textwidth}
    \includegraphics[width=\textwidth]{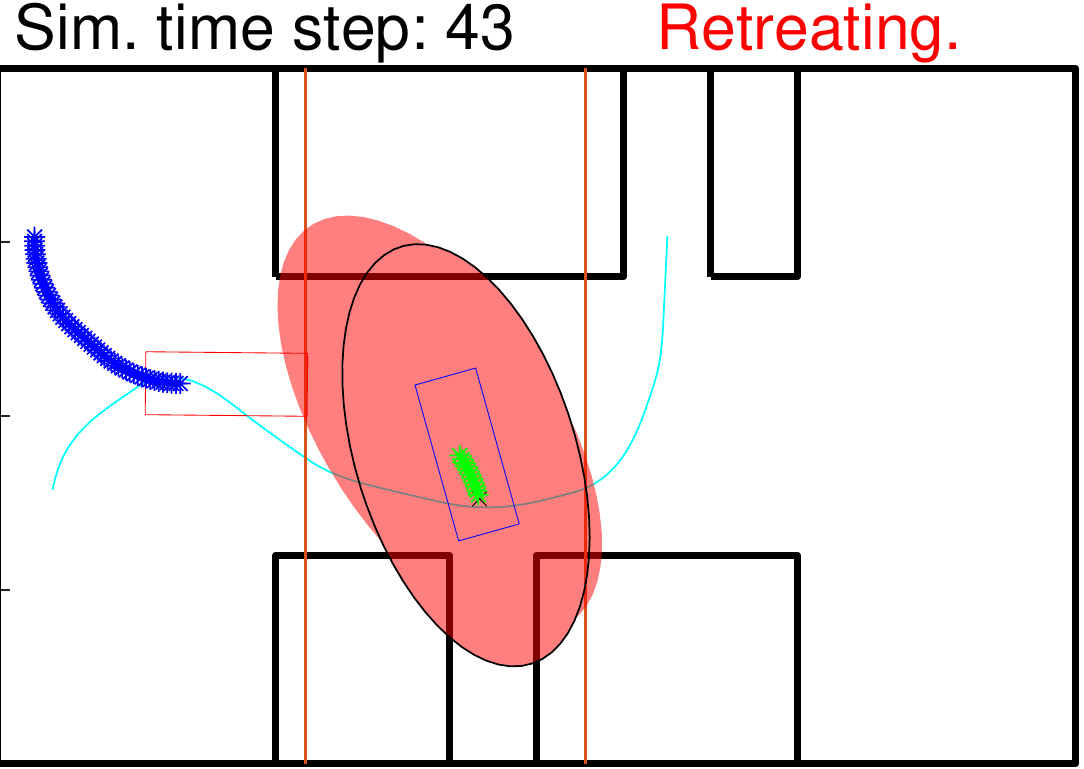}
    \captionof*{figure}{(b)}
   \end{minipage} &
   \begin{minipage}{.23\textwidth}
    \includegraphics[width=\textwidth]{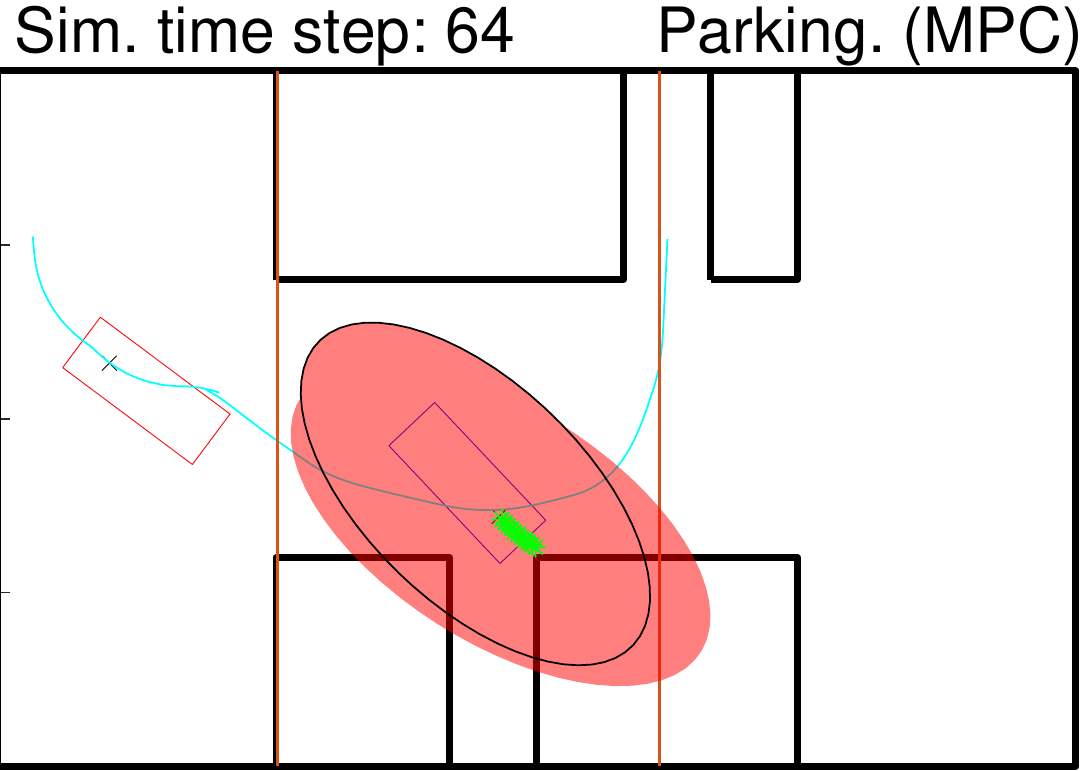}
    \captionof*{figure}{(c)}
   \end{minipage} &
    \begin{minipage}{.23\textwidth}
    \includegraphics[width=\textwidth]{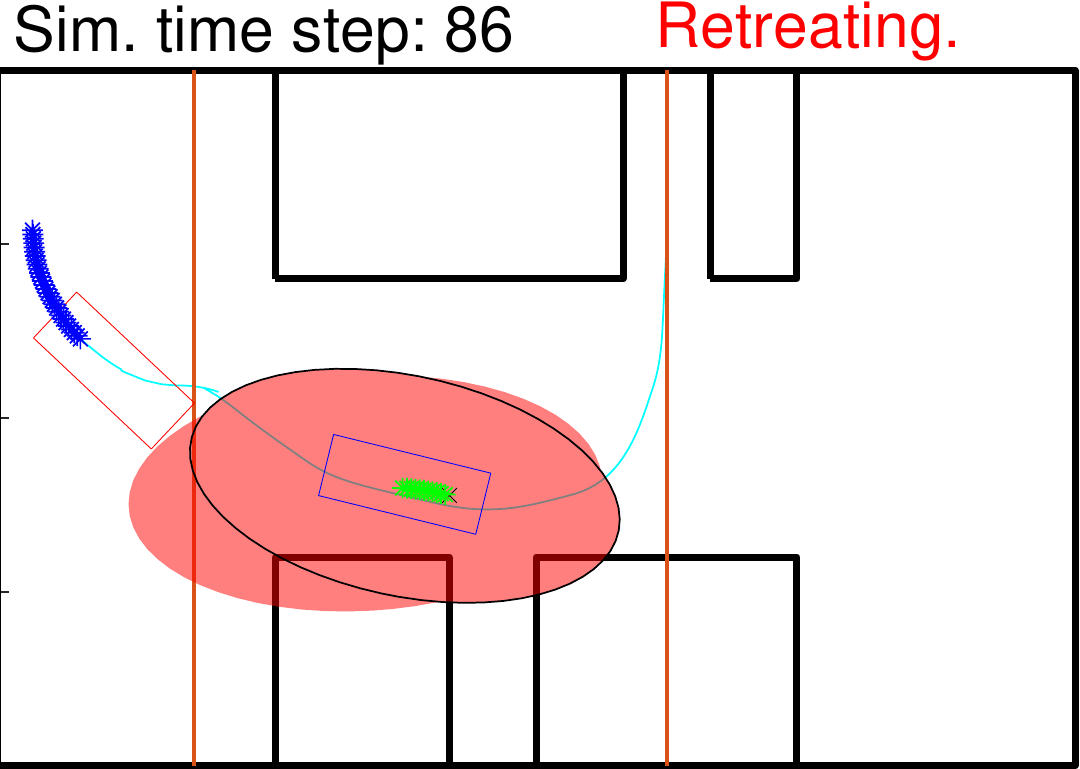}
    \captionof*{figure}{(d)}
   \end{minipage}\\
   \begin{minipage}{.23\textwidth}
    \includegraphics[width=\textwidth]{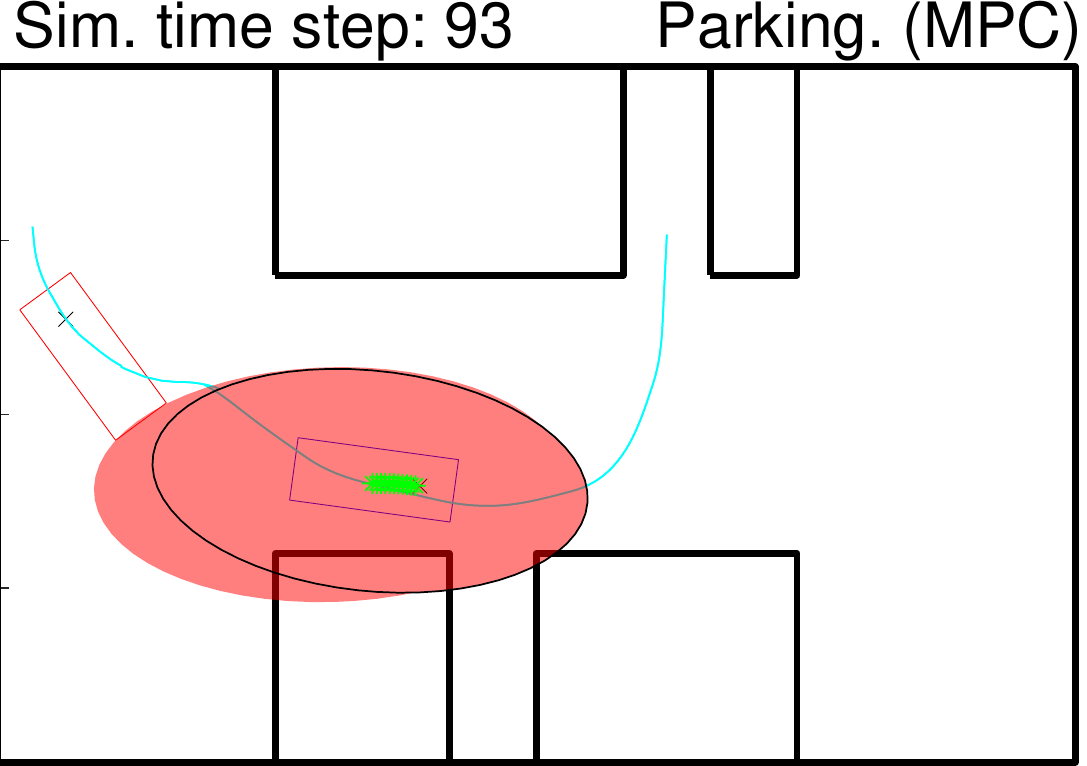}
    \captionof*{figure}{(e)}
   \end{minipage} &
    \begin{minipage}{.23\textwidth}
    \includegraphics[width=\textwidth]{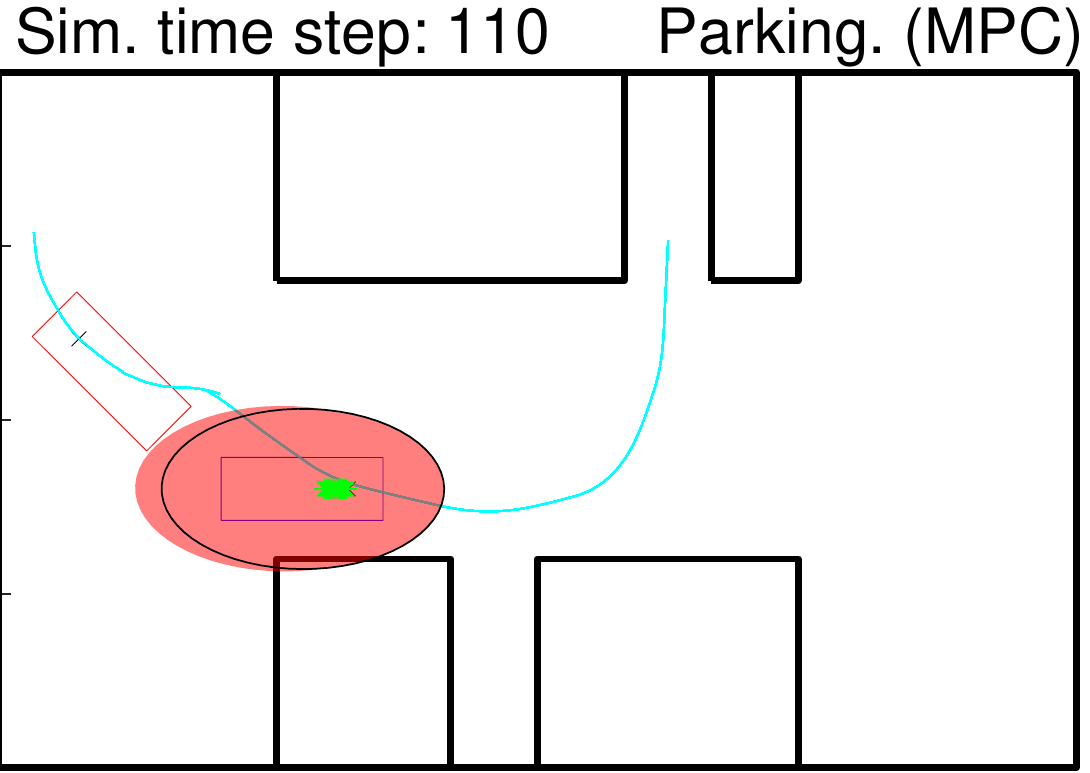}
    \captionof*{figure}{(f)}
   \end{minipage} &
   \begin{minipage}{.23\textwidth}
    \includegraphics[width=\textwidth]{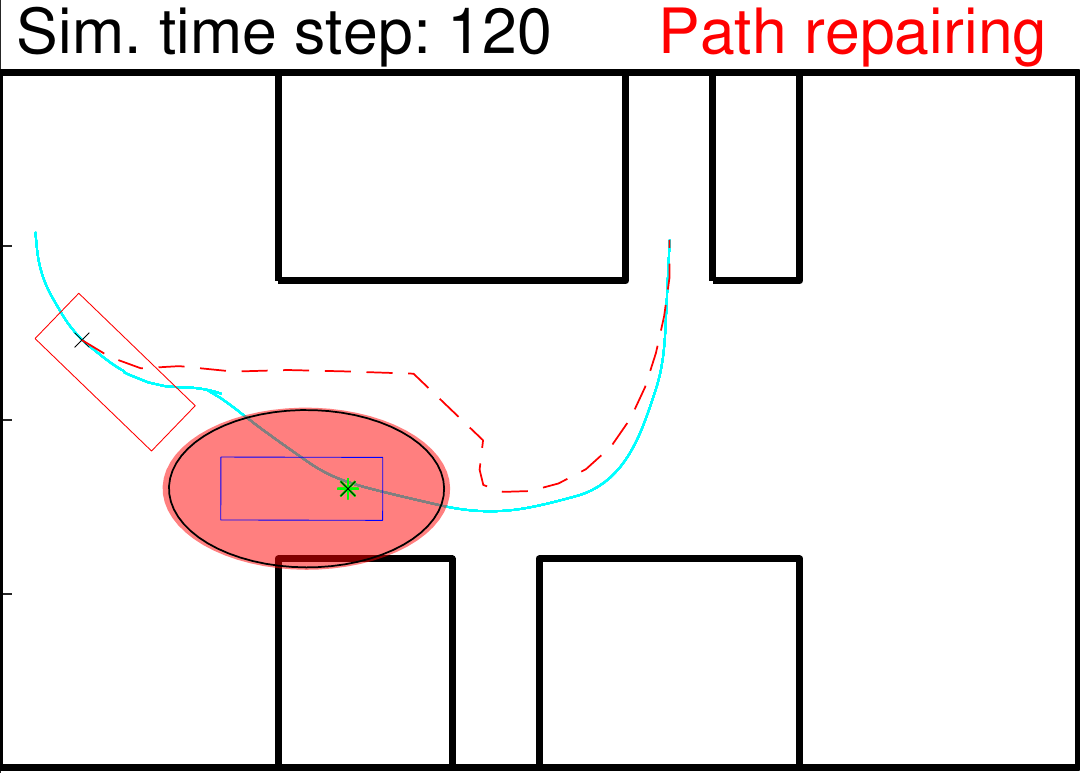}
    \captionof*{figure}{(g)}
   \end{minipage} &
    \begin{minipage}{.23\textwidth}
    \includegraphics[width=\textwidth]{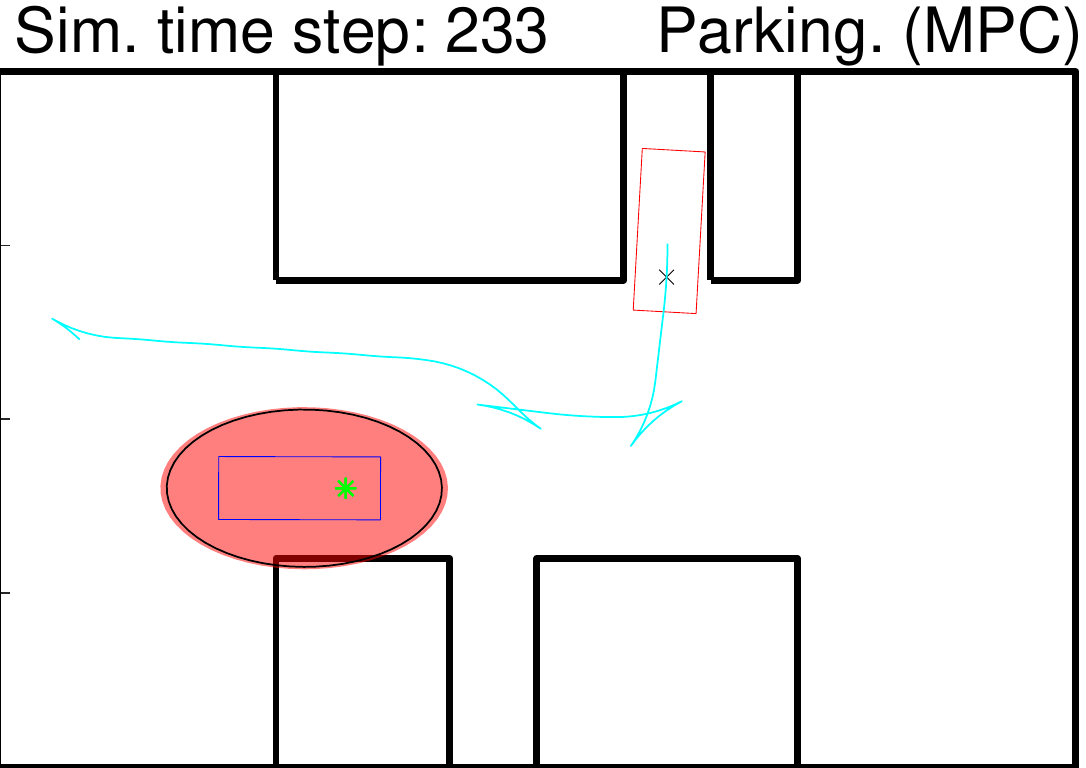}
    \captionof*{figure}{(h)}
   \end{minipage}
  \end{tabular}
    \caption{Simulation results in one of the parking scenarios. (0.25 s/time step)}
  \label{fig:sim}
\end{figure*}

\begin{figure}
	\centering
    \includegraphics[width=0.9\linewidth]{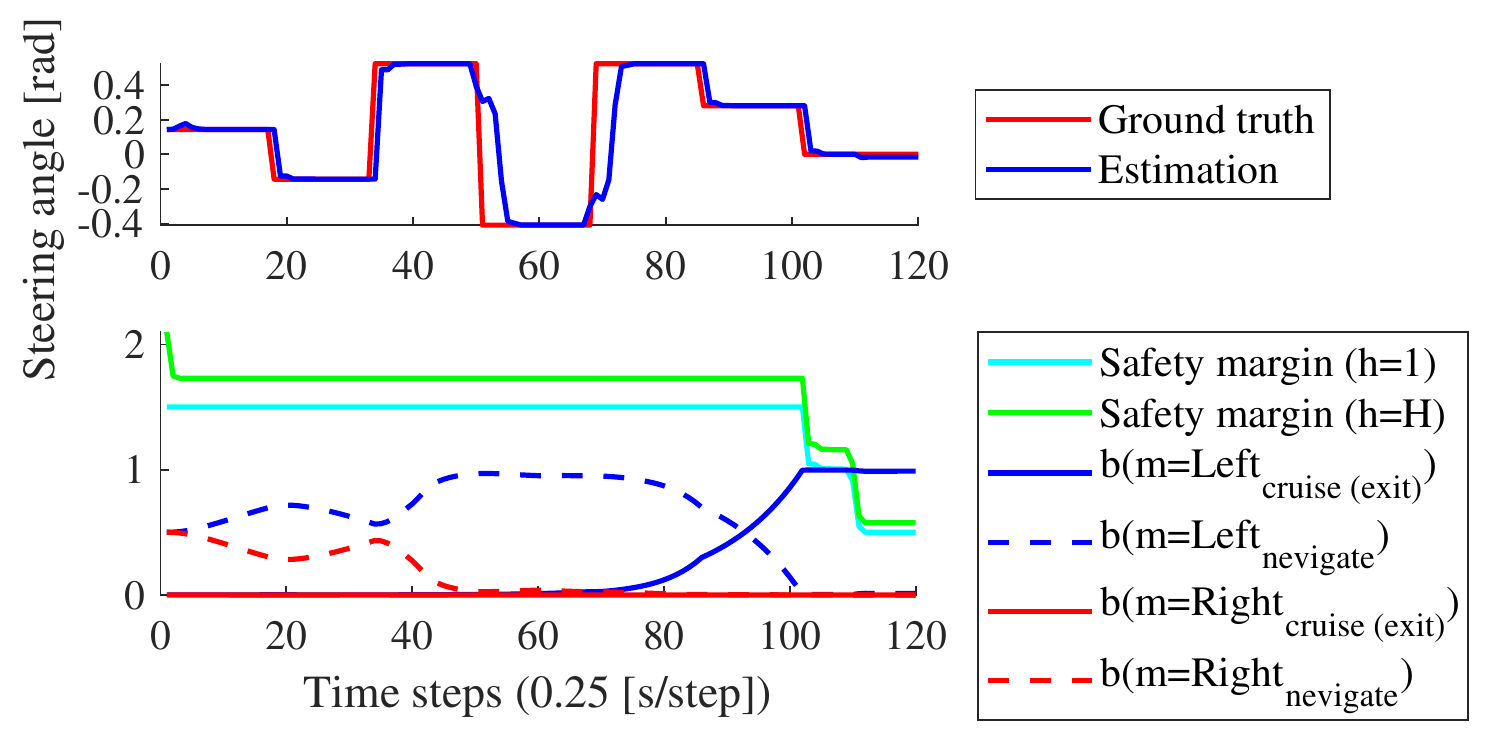}
	\caption{The simulation result of steering estimation, mode estimation, and safety margin.}
	\label{fig:data}
\end{figure}
The proposed system is tested by simulation, which is conducted on a 6-core Intel i7 3.7GHz desktop with Matlab R2020a. The prediction horizon is $10$, i.e., $H=10$. The integrated system is set to runs at a rate of 4 Hz (all calculation in each time step is finished within 0.1 second). Due to space limitation, this section presents one  of the simulation results as an example. More simulation results can be found in the video available at {\tt\small jessicaleu24.github.io/ICRA2022.html}.

As shown in Fig.~\ref{fig:sim}(a), the ego AV (red box) first performs parking by tracking the reference trajectory $\mathcal{P}_{ref}$ (light-blue line) while avoiding collision with the OV (blue box). The safety margins are illustrated by red shaded areas, where the one with black edge is for the current time step and the one without edge is for the $Hth$ future time step (the value is shown in Fig.~\ref{fig:data}). As the OV comes out from its parking spot, it moves towards the ego AV. The ego AV needs to retreat temporarily to make space for the OV (Fig.~\ref{fig:sim}(b)$\sim$(d)). Although the ego AV might be able to avoid collision by backing up along the original reference, this movement may be dangerous since that is the direction where the OV is heading towards and will potentially block the OV. This showcases the importance of considering the long-term mode to construct a collision field (as shown in Fig.~\ref{fig:retreat}) of the OV in retreat planning. In Fig.~\ref{fig:sim}(b)-\ref{fig:sim}(c), the retreat planner is activated to update the $\mathcal{P}_{ref}$. When violation happens again (Fig.~\ref{fig:sim}(d)), the retreat planner updates $\mathcal{P}_{ref}$ once more. As the OV completely leaves the parking spot, the estimator detects the mode switch (from "maneuver left" to "cruise (exit) left" in Fig.~\ref{fig:data}), therefore the safety bound is removed (Fig.~\ref{fig:sim}(e)) and the safety controller continues to follow the reference path (Fig.~\ref{fig:sim}(e)-\ref{fig:sim}(f)). Later, the OV stops at the entrance of the road and blocks the original reference trajectory of the ego AV. Therefore, the ego AV tries repairing the trajectory (Fig.~\ref{fig:sim}(g)). However, the repairing fails (the solver converges to an infeasible red path). Because of the narrow space, cusps are required in the maneuver, which is generally hard for an optimization-based planner to generate. Therefore, a new trajectory from the central control is required. Once the updated trajectory $\mathcal{P}_{ref}$ (Fig.~\ref{fig:sim}(h)) is received, the safety controller will start following it until the ego AV reaches the goal. In the video, we show a successful path repair in demo 1 and the effectiveness of the safety bound (so that the retreat planner won't be triggered unnecessarily) in demo 3.

\section{Conclusion and Future Work}\label{sec:co}

This paper presents an integrated motion planning strategy for an AV to park in dynamic environments. A hybrid environment predictor incorporates the model-based short-term motion prediction and a driver behavior cost-map to make long-term prediction of an OV. A strategic motion planner, composed of an MPC-based safety controller, a search-based retreating planner, and an optimization-based repairing planner, strikes a good balance between safety, plan feasibility, and smooth maneuver by leveraging the advantages of optimization-based and search-based ideas. Depending on the predictor and the AV's objective, the strategic motion planner generates safe and smooth trajectories that bring the AV to the target directly or through an intermediate safe spot to yield to OVs. Simulation results demonstrate that the proposed approach enables the ego AV to plan safely and move smoothly in complicated dynamic parking environments. Future work includes: 1) perform rigorous analysis of the integration strategy, e.g., formal method analysis; 2) generalize and verify the proposed strategy in multi-OVs environment; 3) perform real-time simulation in platforms, for instance dSPACE, for comprehensive assessment of performance and computation load.

\section{Acknowledgement}

The authors would like to thank Karl Berntop, Rien Quirynen, Marcel Menner, and Abraham P. Vinod for their valuable comments. 
\balance

\bibliographystyle{IEEEtran}
\bibliography{PathPlanning}

\begin{thebibliography}{10}
\providecommand{\url}[1]{#1}
\csname url@samestyle\endcsname
\providecommand{\newblock}{\relax}
\providecommand{\bibinfo}[2]{#2}
\providecommand{\BIBentrySTDinterwordspacing}{\spaceskip=0pt\relax}
\providecommand{\BIBentryALTinterwordstretchfactor}{4}
\providecommand{\BIBentryALTinterwordspacing}{\spaceskip=\fontdimen2\font plus
\BIBentryALTinterwordstretchfactor\fontdimen3\font minus
  \fontdimen4\font\relax}
\providecommand{\BIBforeignlanguage}[2]{{%
\expandafter\ifx\csname l@#1\endcsname\relax
\typeout{** WARNING: IEEEtran.bst: No hyphenation pattern has been}%
\typeout{** loaded for the language `#1'. Using the pattern for}%
\typeout{** the default language instead.}%
\else
\language=\csname l@#1\endcsname
\fi
#2}}
\providecommand{\BIBdecl}{\relax}
\BIBdecl

\bibitem{conner2007valet}
D.~C. Conner, H.~Kress-Gazit, H.~Choset, A.~A. Rizzi, and G.~J. Pappas, ``Valet
  parking without a valet,'' in \emph{2007 IEEE/RSJ international conference on
  intelligent robots and systems}.\hskip 1em plus 0.5em minus 0.4em\relax IEEE,
  2007, pp. 572--577.

\bibitem{min2013design}
K.-W. Min and J.-D. Choi, ``Design and implementation of autonomous vehicle
  valet parking system,'' in \emph{16th International IEEE Conference on
  Intelligent Transportation Systems (ITSC 2013)}.\hskip 1em plus 0.5em minus
  0.4em\relax IEEE, 2013, pp. 2082--2087.

\bibitem{leu2019motion}
J.~Leu and M.~Tomizuka, ``Motion planning for industrial mobile robots with
  closed-loop stability enhanced prediction,'' in \emph{Dynamic Systems and
  Control Conference}, vol. 59162.\hskip 1em plus 0.5em minus 0.4em\relax
  American Society of Mechanical Engineers, 2019, p. V003T19A009.

\bibitem{costmap08}
D.~Ferguson, T.~M. Howard, and M.~Likhachev, ``Motion planning in urban
  environments: Part ii,'' in \emph{2008 IEEE/RSJ International Conference on
  Intelligent Robots and Systems}, 2008, pp. 1070--1076.

\bibitem{costmap14}
T.~Howard, M.~Pivtoraiko, R.~A. Knepper, and A.~Kelly, ``Model-predictive
  motion planning: Several key developments for autonomous mobile robots,''
  \emph{IEEE Robotics Automation Magazine}, vol.~21, no.~1, pp. 64--73, 2014.

\bibitem{dolgov2008practical}
D.~Dolgov, S.~Thrun, M.~Montemerlo, and J.~Diebel, ``Practical search
  techniques in path planning for autonomous driving,'' \emph{Ann Arbor}, vol.
  1001, no. 48105, pp. 18--80, 2008.

\bibitem{mcnaughton2011motion}
M.~McNaughton, C.~Urmson, J.~M. Dolan, and J.-W. Lee, ``Motion planning for
  autonomous driving with a conformal spatiotemporal lattice,'' in \emph{2011
  IEEE International Conference on Robotics and Automation}.\hskip 1em plus
  0.5em minus 0.4em\relax IEEE, 2011, pp. 4889--4895.

\bibitem{ma2015efficient}
L.~Ma, J.~Xue, K.~Kawabata, J.~Zhu, C.~Ma, and N.~Zheng, ``Efficient
  sampling-based motion planning for on-road autonomous driving,'' \emph{IEEE
  Transactions on Intelligent Transportation Systems}, vol.~16, no.~4, pp.
  1961--1976, 2015.

\bibitem{islam2016connect}
F.~Islam, V.~Narayanan, and M.~Likhachev, ``A*-connect: Bounded suboptimal
  bidirectional heuristic search,'' in \emph{2016 IEEE International Conference
  On Robotics and Automation (ICRA)}.\hskip 1em plus 0.5em minus 0.4em\relax
  IEEE, 2016, pp. 2752--2758.

\bibitem{aine2016multi}
S.~Aine, S.~Swaminathan, V.~Narayanan, V.~Hwang, and M.~Likhachev,
  ``Multi-heuristic a,'' \emph{The International Journal of Robotics Research},
  vol.~35, no. 1-3, pp. 224--243, 2016.

\bibitem{chen2017constrained}
J.~Chen, W.~Zhan, and M.~Tomizuka, ``Constrained iterative lqr for on-road
  autonomous driving motion planning,'' in \emph{2017 IEEE 20th International
  Conference on Intelligent Transportation Systems (ITSC)}.\hskip 1em plus
  0.5em minus 0.4em\relax IEEE, 2017, pp. 1--7.

\bibitem{hsieh2008parking}
M.~F. Hsieh and U.~Ozguner, ``A parking algorithm for an autonomous vehicle,''
  in \emph{2008 IEEE Intelligent Vehicles Symposium}.\hskip 1em plus 0.5em
  minus 0.4em\relax IEEE, 2008, pp. 1155--1160.

\bibitem{kummerle2009autonomous}
R.~Kummerle, D.~Hahnel, D.~Dolgov, S.~Thrun, and W.~Burgard, ``Autonomous
  driving in a multi-level parking structure,'' in \emph{2009 IEEE
  International Conference on Robotics and Automation}.\hskip 1em plus 0.5em
  minus 0.4em\relax IEEE, 2009, pp. 3395--3400.

\bibitem{han2011unified}
L.~Han, Q.~H. Do, and S.~Mita, ``Unified path planner for parking an autonomous
  vehicle based on rrt,'' in \emph{2011 IEEE International Conference on
  Robotics and Automation}.\hskip 1em plus 0.5em minus 0.4em\relax IEEE, 2011,
  pp. 5622--5627.

\bibitem{klemm2016autonomous}
S.~Klemm, M.~Essinger, J.~Oberl{\"a}nder, M.~R. Zofka, F.~Kuhnt, M.~Weber,
  R.~Kohlhaas, A.~Kohs, A.~Roennau, T.~Schamm \emph{et~al.}, ``Autonomous
  multi-story navigation for valet parking,'' in \emph{2016 IEEE 19th
  International Conference on Intelligent Transportation Systems (ITSC)}.\hskip
  1em plus 0.5em minus 0.4em\relax IEEE, 2016, pp. 1126--1133.

\bibitem{tazaki2017parking}
Y.~Tazaki, H.~Okuda, and T.~Suzuki, ``Parking trajectory planning using
  multiresolution state roadmaps,'' \emph{IEEE Transactions on Intelligent
  Vehicles}, vol.~2, no.~4, pp. 298--307, 2017.

\bibitem{wang2019improved}
Y.~Wang, ``Improved a-search guided tree construction for kinodynamic
  planning,'' in \emph{2019 International Conference on Robotics and Automation
  (ICRA)}.\hskip 1em plus 0.5em minus 0.4em\relax IEEE, 2019, pp. 5530--5536.

\bibitem{rawlings2017model}
J.~B. Rawlings, D.~Q. Mayne, and M.~Diehl, \emph{Model predictive control:
  theory, computation, and design}.\hskip 1em plus 0.5em minus 0.4em\relax Nob
  Hill Publishing Madison, WI, 2017, vol.~2.

\bibitem{schubert2008comparison}
R.~Schubert, E.~Richter, and G.~Wanielik, ``Comparison and evaluation of
  advanced motion models for vehicle tracking,'' in \emph{2008 11th
  international conference on information fusion}.\hskip 1em plus 0.5em minus
  0.4em\relax IEEE, 2008, pp. 1--6.

\bibitem{houenou2013vehicle}
A.~Houenou, P.~Bonnifait, V.~Cherfaoui, and W.~Yao, ``Vehicle trajectory
  prediction based on motion model and maneuver recognition,'' in \emph{2013
  IEEE/RSJ international conference on intelligent robots and systems}.\hskip
  1em plus 0.5em minus 0.4em\relax IEEE, 2013, pp. 4363--4369.

\bibitem{4895669}
M.~Althoff, O.~Stursberg, and M.~Buss, ``Model-based probabilistic collision
  detection in autonomous driving,'' \emph{IEEE Transactions on Intelligent
  Transportation Systems}, vol.~10, no.~2, pp. 299--310, 2009.

\bibitem{shen2020collision}
X.~Shen, E.~L. Zhu, Y.~R. St{\"u}rz, and F.~Borrelli, ``Collision avoidance in
  tightly-constrained environments without coordination: a hierarchical control
  approach,'' \emph{arXiv preprint arXiv:2011.00413}, 2020.

\bibitem{2015classify}
J.~Schlechtriemen, F.~Wirthmueller, A.~Wedel, G.~Breuel, and K.-D. Kuhnert,
  ``When will it change the lane? a probabilistic regression approach for
  rarely occurring events,'' in \emph{2015 IEEE Intelligent Vehicles Symposium
  (IV)}, 2015, pp. 1373--1379.

\bibitem{shen2020parkpredict}
X.~Shen, I.~Batkovic, V.~Govindarajan, P.~Falcone, T.~Darrell, and F.~Borrelli,
  ``Parkpredict: Motion and intent prediction of vehicles in parking lots,'' in
  \emph{2020 IEEE Intelligent Vehicles Symposium (IV)}.\hskip 1em plus 0.5em
  minus 0.4em\relax IEEE, 2020, pp. 1170--1175.

\bibitem{deo2018would}
N.~Deo, A.~Rangesh, and M.~M. Trivedi, ``How would surround vehicles move? a
  unified framework for maneuver classification and motion prediction,''
  \emph{IEEE Transactions on Intelligent Vehicles}, vol.~3, no.~2, pp.
  129--140, 2018.

\bibitem{6025208}
C.~Laugier, I.~E. Paromtchik, M.~Perrollaz, M.~Yong, J.-D. Yoder, C.~Tay,
  K.~Mekhnacha, and A.~Nègre, ``Probabilistic analysis of dynamic scenes and
  collision risks assessment to improve driving safety,'' \emph{IEEE
  Intelligent Transportation Systems Magazine}, vol.~3, no.~4, pp. 4--19, 2011.

\bibitem{lefevre2014survey}
S.~Lef{\`e}vre, D.~Vasquez, and C.~Laugier, ``A survey on motion prediction and
  risk assessment for intelligent vehicles,'' \emph{ROBOMECH journal}, vol.~1,
  no.~1, pp. 1--14, 2014.

\bibitem{jeong2018sampling}
Y.~Jeong, S.~Kim, B.~R. Jo, H.~Shin, and K.~Yi, ``Sampling based vehicle motion
  planning for autonomous valet parking with moving obstacles,''
  \emph{International Journal of Automotive Engineering}, vol.~9, no.~4, pp.
  215--222, 2018.

\bibitem{govea2004moving}
D.~A.~V. Govea, F.~Large, T.~Fraichard, and C.~Laugier, ``Moving obstacles'
  motion prediction for autonomous navigation,'' in \emph{Proc. of the Int.
  Conf. on Control, Automation, Robotics and Vision}, 2004.

\bibitem{HarNilRap68}
P.~E. Hart, N.~J. Nilsson, and B.~Raphael, ``A formal basis for the heuristic
  determination of minimum cost paths,'' \emph{IEEE Transactions on Systems
  Science and Cybernetics}, vol.~4, no.~2, pp. 100--107, 1968.

\bibitem{Ste93}
A.~Stentz, ``Optimal and efficient path planning for unknown and dynamic
  environments,'' Robotics Institute, Carnegie Mellon University, Tech. Rep.
  CMU-RI-TR-93-20, 1993.

\bibitem{LikFerGor05}
M.~Likhachev, D.~I. Ferguson, G.~J. Gordon, A.~Stentz, and S.~Thrun, ``{Anytime
  Dynamic A*}: An anytime, replanning algorithm,'' in \emph{Proc. 2005 ICAPS},
  2005, pp. 262--271.

\bibitem{kavraki1996probabilistic}
L.~E. Kavraki, P.~Svestka, J.-C. Latombe, and M.~H. Overmars, ``Probabilistic
  roadmaps for path planning in high-dimensional configuration spaces,''
  \emph{IEEE transactions on Robotics and Automation}, vol.~12, no.~4, pp.
  566--580, 1996.

\bibitem{lavalle2001randomized}
S.~M. LaValle and J.~J. Kuffner~Jr, ``Randomized kinodynamic planning,''
  \emph{The international journal of robotics research}, vol.~20, no.~5, pp.
  378--400, 2001.

\bibitem{karaman2011sampling}
S.~Karaman and E.~Frazzoli, ``Sampling-based algorithms for optimal motion
  planning,'' \emph{The international journal of robotics research}, vol.~30,
  no.~7, pp. 846--894, 2011.

\bibitem{zucker2013chomp}
M.~Zucker, N.~Ratliff, A.~D. Dragan, M.~Pivtoraiko, M.~Klingensmith, C.~M.
  Dellin, J.~A. Bagnell, and S.~S. Srinivasa, ``Chomp: Covariant hamiltonian
  optimization for motion planning,'' \emph{The International Journal of
  Robotics Research}, vol.~32, no. 9-10, pp. 1164--1193, 2013.

\bibitem{schulman2014motion}
J.~Schulman, Y.~Duan, J.~Ho, A.~Lee, I.~Awwal, H.~Bradlow, J.~Pan, S.~Patil,
  K.~Goldberg, and P.~Abbeel, ``Motion planning with sequential convex
  optimization and convex collision checking,'' \emph{The International Journal
  of Robotics Research}, vol.~33, no.~9, pp. 1251--1270, 2014.

\bibitem{gutjahr2016lateral}
B.~Gutjahr, L.~Gr{\"o}ll, and M.~Werling, ``Lateral vehicle trajectory
  optimization using constrained linear time-varying mpc,'' \emph{IEEE
  Transactions on Intelligent Transportation Systems}, vol.~18, no.~6, pp.
  1586--1595, 2016.

\bibitem{zhang2018autonomous}
X.~Zhang, A.~Liniger, A.~Sakai, and F.~Borrelli, ``Autonomous parking using
  optimization-based collision avoidance,'' in \emph{2018 IEEE Conference on
  Decision and Control (CDC)}.\hskip 1em plus 0.5em minus 0.4em\relax IEEE,
  2018, pp. 4327--4332.

\bibitem{zucker2010optimization}
M.~Zucker, J.~A. Bagnell, C.~G. Atkeson, and J.~Kuffner, ``An optimization
  approach to rough terrain locomotion,'' in \emph{2010 IEEE International
  Conference on Robotics and Automation}.\hskip 1em plus 0.5em minus
  0.4em\relax IEEE, 2010, pp. 3589--3595.

\bibitem{xu2012real}
W.~Xu, J.~Wei, J.~M. Dolan, H.~Zhao, and H.~Zha, ``A real-time motion planner
  with trajectory optimization for autonomous vehicles,'' in \emph{2012 IEEE
  International Conference on Robotics and Automation}.\hskip 1em plus 0.5em
  minus 0.4em\relax IEEE, 2012, pp. 2061--2067.

\bibitem{dai2018improving}
S.~Dai, M.~Orton, S.~Schaffert, A.~Hofmann, and B.~Williams, ``Improving
  trajectory optimization using a roadmap framework,'' in \emph{2018 IEEE/RSJ
  International Conference on Intelligent Robots and Systems (IROS)}.\hskip 1em
  plus 0.5em minus 0.4em\relax IEEE, 2018, pp. 8674--8681.

\bibitem{leu2021efficient}
J.~Leu, G.~Zhang, L.~Sun, and M.~Tomizuka, ``Efficient robot motion planning
  via sampling and optimization,'' in \emph{2021 American Control Conference
  (ACC)}.\hskip 1em plus 0.5em minus 0.4em\relax IEEE, 2021, pp. 4196--4202.

\bibitem{UrmAnhBag08}
C.~Urmson, J.~Anhalt, D.~Bagnell, C.~Baker, B.~Bittner, M.~N. Clark, J.~Dolan,
  D.~Duggins, T.~Galatali, C.~Geyer \emph{et~al.}, ``Autonomous driving in
  urban environments: Boss and the urban challenge,'' \emph{Journal of Field
  Robotics}, vol.~25, no.~8, pp. 425--466, 2008.

\bibitem{montemerlo2008junior}
M.~Montemerlo, J.~Becker, S.~Bhat, H.~Dahlkamp, D.~Dolgov, S.~Ettinger,
  D.~Haehnel, T.~Hilden, G.~Hoffmann, B.~Huhnke \emph{et~al.}, ``Junior: The
  stanford entry in the urban challenge,'' \emph{Journal of field Robotics},
  vol.~25, no.~9, pp. 569--597, 2008.

\bibitem{dolgov2010path}
D.~Dolgov, S.~Thrun, M.~Montemerlo, and J.~Diebel, ``Path planning for
  autonomous vehicles in unknown semi-structured environments,'' \emph{The
  International Journal of Robotics Research}, vol.~29, no.~5, pp. 485--501,
  2010.

\bibitem{chen2015kinodynamic}
C.~Chen, M.~Rickert, and A.~Knoll, ``Kinodynamic motion planning with
  space-time exploration guided heuristic search for car-like robots in dynamic
  environments,'' in \emph{2015 IEEE/RSJ International Conference on
  Intelligent Robots and Systems (IROS)}.\hskip 1em plus 0.5em minus
  0.4em\relax IEEE, 2015, pp. 2666--2671.

\bibitem{ajanovic2018search}
Z.~Ajanovic, B.~Lacevic, B.~Shyrokau, M.~Stolz, and M.~Horn, ``Search-based
  optimal motion planning for automated driving,'' in \emph{2018 IEEE/RSJ
  International Conference on Intelligent Robots and Systems (IROS)}.\hskip 1em
  plus 0.5em minus 0.4em\relax IEEE, 2018, pp. 4523--4530.

\bibitem{chen2015path}
C.~Chen, M.~Rickert, and A.~Knoll, ``Path planning with orientation-aware space
  exploration guided heuristic search for autonomous parking and maneuvering,''
  in \emph{2015 IEEE Intelligent Vehicles Symposium (IV)}.\hskip 1em plus 0.5em
  minus 0.4em\relax IEEE, 2015, pp. 1148--1153.

\bibitem{klaudt2017priori}
S.~Klaudt, A.~Zlocki, and L.~Eckstein, ``A-priori map information and path
  planning for automated valet-parking,'' in \emph{2017 IEEE Intelligent
  Vehicles Symposium (IV)}.\hskip 1em plus 0.5em minus 0.4em\relax IEEE, 2017,
  pp. 1770--1775.

\bibitem{koenig2002d}
S.~Koenig and M.~Likhachev, ``D\^{}* lite,'' \emph{AAAI}, vol.~15, 2002.

\bibitem{sun2010moving}
X.~Sun, W.~Yeoh, and S.~Koenig, ``Moving target d* lite,'' in \emph{Proceedings
  of the 9th International Conference on Autonomous Agents and Multiagent
  Systems: volume 1-Volume 1}, 2010, pp. 67--74.

\bibitem{oral2015mod}
T.~Oral and F.~Polat, ``Mod* lite: an incremental path planning algorithm
  taking care of multiple objectives,'' \emph{IEEE Transactions on
  Cybernetics}, vol.~46, no.~1, pp. 245--257, 2015.

\bibitem{ferguson2006replanning}
D.~Ferguson, N.~Kalra, and A.~Stentz, ``Replanning with rrts,'' in
  \emph{Proceedings 2006 IEEE International Conference on Robotics and
  Automation, 2006. ICRA 2006.}\hskip 1em plus 0.5em minus 0.4em\relax IEEE,
  2006, pp. 1243--1248.

\bibitem{chandler2017online}
B.~Chandler and M.~A. Goodrich, ``Online rrt* and online fmt*: Rapid replanning
  with dynamic cost,'' in \emph{2017 IEEE/RSJ International Conference on
  Intelligent Robots and Systems (IROS)}.\hskip 1em plus 0.5em minus
  0.4em\relax IEEE, 2017, pp. 6313--6318.

\bibitem{adiyatov2017novel}
O.~Adiyatov and H.~A. Varol, ``A novel rrt*-based algorithm for motion planning
  in dynamic environments,'' in \emph{2017 IEEE International Conference on
  Mechatronics and Automation (ICMA)}.\hskip 1em plus 0.5em minus 0.4em\relax
  IEEE, 2017, pp. 1416--1421.

\bibitem{qi2020mod}
J.~Qi, H.~Yang, and H.~Sun, ``Mod-rrt*: A sampling-based algorithm for robot
  path planning in dynamic environment,'' \emph{IEEE Transactions on Industrial
  Electronics}, 2020.

\bibitem{ding2018trajectory}
W.~Ding, W.~Gao, K.~Wang, and S.~Shen, ``Trajectory replanning for quadrotors
  using kinodynamic search and elastic optimization,'' in \emph{2018 IEEE
  International Conference on Robotics and Automation (ICRA)}.\hskip 1em plus
  0.5em minus 0.4em\relax IEEE, 2018, pp. 7595--7602.

\bibitem{ding2019efficient}
------, ``An efficient b-spline-based kinodynamic replanning framework for
  quadrotors,'' \emph{IEEE Transactions on Robotics}, vol.~35, no.~6, pp.
  1287--1306, 2019.

\bibitem{9062306}
X.~Zhang, A.~Liniger, and F.~Borrelli, ``Optimization-based collision
  avoidance,'' \emph{IEEE Transactions on Control Systems Technology}, vol.~29,
  no.~3, pp. 972--983, 2021.

\bibitem{9183957}
J.-H. Jhang and F.-L. Lian, ``An autonomous parking system of optimally
  integrating bidirectional rapidly-exploring random trees* and
  parking-oriented model predictive control,'' \emph{IEEE Access}, vol.~8, pp.
  163\,502--163\,523, 2020.

\bibitem{2020parkingsystem}
C.~Jang, C.~Kim, S.~Lee, S.~Kim, S.~Lee, and M.~Sunwoo, ``Re-plannable
  automated parking system with a standalone around view monitor for narrow
  parking lots,'' \emph{IEEE Transactions on Intelligent Transportation
  Systems}, vol.~21, no.~2, pp. 777--790, 2020.

\bibitem{HouPat98}
M.~Hou and R.~Patton, ``Input observability and input reconstruction,''
  vol.~34, no.~6, pp. 789--794, Jun. 1998.

\bibitem{dailong}
S.~Dai and Y.~Wang, ``Long-horizon motion planning for autonomous vehicle
  parking incorporating incomplete map information.''

\bibitem{cpb}
C.~L. Baker, R.~Saxe, and J.~B. Tenenbaum, ``Action understanding as inverse
  planning,'' \emph{Cognition}, vol. 113, no.~3, pp. 329--349, 2009.

\bibitem{AndGilHor19}
J.~A.~E. Andersson, J.~Gillis, G.~Horn, J.~B. Rawlings, and M.~Diehl,
  ``{CasADi} -- {A} software framework for nonlinear optimization and optimal
  control,'' \emph{Mathematical Programming Computation}, vol.~11, no.~1, pp.
  1--36, 2019.

\end{thebibliography}
\end{document}